\documentclass{article}

\usepackage{microtype}
\usepackage{graphicx}
\graphicspath{ {./} }
\usepackage{subfigure}
\usepackage{booktabs} 
\usepackage{multirow}
\usepackage{array}
\usepackage{wrapfig}
\usepackage{siunitx}
\usepackage{float}
\usepackage{bbold}
\usepackage{textcomp}
\usepackage{csvsimple}
\csvstyle{mystyle}{
    table head=\midrule,
    no head,
    late after line=\\,
    late after first line=\\\midrule,
    }
\newcommand{\respectpercent}{\catcode`\%=12\relax}
\newcommand{\beginsupplement}{%
    \setcounter{section}{0}
    \renewcommand{\thesection}{S\arabic{section}}%
    \setcounter{table}{0}
    \renewcommand{\thetable}{S\arabic{table}}%
    \setcounter{figure}{0}
    \renewcommand{\thefigure}{S\arabic{figure}}%
    }

\usepackage{hyperref}

\usepackage[accepted]{icml2020_copy}

\begin{document}

\twocolumn[
\icmltitle{Graph Neural Networks for the Prediction of Substrate-Specific Organic Reaction Conditions}



\icmlsetsymbol{equal}{*}

\begin{icmlauthorlist}
\icmlauthor{Serim Ryou}{equal,cvl}
\icmlauthor{Michael R. Maser}{equal,cce}
\icmlauthor{Alexander Y. Cui}{equal,cms}
\icmlauthor{Travis J. DeLano}{cce}
\icmlauthor{Yisong Yue}{cms}
\icmlauthor{Sarah E. Reisman}{cce}
\end{icmlauthorlist}

\icmlaffiliation{cce}{Division of Chemistry and Chemical Engineering, California Institute of Technology, Pasadena, California, USA}
\icmlaffiliation{cvl}{Computational Vision Lab, California Institute of Technology, Pasadena, California, USA}
\icmlaffiliation{cms}{Department of Computing and Mathematical Sciences, California Institute of Technology, Pasadena, California, USA}

\icmlcorrespondingauthor{Michael R. Maser}{mmaser@caltech.edu}
\icmlcorrespondingauthor{Sarah E. Reisman}{reisman@caltech.edu}

\icmlkeywords{Graph Neural Networks, Machine Learning, Relational Networks, Organic Chemistry, Cross-Coupling Reaction, Predicting Experimental Conditions}

\vskip 0.3in
]



\printAffiliationsAndNotice{\icmlEqualContribution} 

\begin{abstract}
We present a systematic investigation using graph neural networks (GNNs) to model organic chemical reactions. To do so, we prepared a dataset collection of four ubiquitous reactions from the organic chemistry literature. We evaluate seven different GNN architectures for classification tasks pertaining to the identification of experimental reagents and conditions. We find that models are able to identify specific graph features that affect reaction conditions and lead to accurate predictions. The results herein show great promise in advancing molecular machine learning.
\end{abstract}
\section{Introduction}
\label{sec:intro}
Graph neural networks (GNNs) have rapidly emerged as powerful predictive tools in the chemistry domain \citep{mater_deep_2019}. Significant developments have been made in medicinal chemistry, where predictors of drug physicochemical properties \citep{coley_graph-convolutional_2019, withnall_building_2020} and graph generative models \citep{elton_deep_2019, blaschke_application_2018} are readily available. Several examples have also been reported in organic synthesis, including for the classification of reaction types \citep{schwaller_data-driven_2019}, the prediction of reaction products \citep{skoraczynski_predicting_2017, coley_prediction_2017}, and for retrosynthetic design \citep{segler_planning_2018, coley_machine_2018}. Despite the recent advancements, relatively few studies have been reported for the prediction of reaction conditions, arguably among the most challenging tasks chemists face \citep{coley_prediction_2017}.
\par Current deep neural networks (NNs) rely on multi-million reaction training sets for sufficient data to make predictions in high-dimensional (1,000$^+$) label spaces \citep{gao_using_2018}. This offers flexibility in the reaction types that can be queried and includes a broad condition space from all of organic chemistry. However, given the sparsity of global datasets, reliable predictions are likely only obtained for the most common conditions of each reaction type, regardless of the structural differences between inputs. This poses a severe limitation for catalytic reactions in that the optimal conditions are often highly dependent on substrate structure \citep{mahatthananchai_catalytic_2012}. It is therefore critical that deep networks distinguish between individual graphs of a reaction type when suggesting appropriate conditions to use.
\par To address the current limitations, we approach this prediction problem at the single reaction level. This allows for detailed learning of structure-condition relationships without the need to classify queries by learned reaction rules. We demonstrate the utility of our approach on curated datasets of four valuable reaction types from organic chemistry: Suzuki couplings \citep{miyaura_palladium-catalyzed_1995}, C–N couplings \citep{bariwal_cn_2013}, Negishi couplings \citep{negishi_selective_1977}, and Pauson–Khand reactions (PKRs) \citep{khand_organocobalt_1973}. Our main contributions can be summarized as follows: 
%
\begin{enumerate}
    \item We apply domain expertise in dataset curation and in the construction of the label space.
    \item We conduct a thorough assessment of modern GNN architectures, which, to our knowledge, has not yet been studied for models of chemical reactivity.
    \item We achieve high accuracy in predicting condition vectors for all four datasets using strict evaluation metrics.
    \item We provide an interpretability analysis to show the structural features informing our predictions.
\end{enumerate}
\section{Approach and Related Work}
\label{sec:approach_related}
%
The work presented herein bears greatest similarity to \citet{gao_using_2018}. In this report, a single GNN was trained on 12 million reactions from the full Reaxys®  database \citep{reaxys_reaxys_nodate} for the task of predicting conditions. Product and reaction fingerprints (FPs) were used as inputs, where the latter was defined as the difference between the product and reactant graphs. Predictions were made in sequence for one possible catalyst, two solvents, and two reagents for all samples, regardless of reaction type. Intermediate predictions were concatenated as one-hot vectors with the FP inputs and passed to each subsequent layer, all of which were fully connected. Reasonable accuracies were obtained, though the inclusion of top-10 rankings and ``close match" labels was required in many cases.
%
\par Our approach offers significant advantages over prior art on several accounts. We model focused reaction sets to obtain fine-grained, graph-specific predictions. Our networks take full graphs of all reaction components as inputs to minimize information loss in encoded structures. Our predictions use reaction-specific roles identified directly from dataset analysis to ensure proper chemical context of output vectors. We analyze only top-1 and top-3 predicted rankings to increase the feasibility of testing suggested results experimentally. We explicitly treat accuracies relative to a naive model to provide a rigorous evaluation framework. Even with these strict metrics, we obtain high accuracies through systematic testing of a suite of GNN architectures on each task.

\section{Methods}
\label{sec:methods}


\textbf{Reaction data}. Literature datasets are obtained from the Reaxys® database and are pre-processed to remove incomplete records. A summary of the prepared datasets is included in Table \ref{tab:rxn}. Detailed processing steps and data analysis can be found in the Supplementary Material (SM), including distributions of molecular properties, reaction yields, and reagent frequencies.\footnote{Since Reaxys® is a subscription database, we are not permitted to publish exported data. We have compiled detailed procedures to prepare each dataset such that those with access can replicate our results. Most academic institutions have full-access Reaxys® subscriptions. We make full reaction label dictionaries and all modeling code available at \url{https://github.com/slryou41/reaction-gcnn}.} A general workflow for dataset preparation is as follows:
\begin{enumerate}
	\item From Reaxys® exports, SMILES string encodings \citep{weininger_smiles_1988} of reactants and products are extracted for each data point.
	\item Full condition vectors including reagents, catalysts, solvents, temperatures, etc. are extracted for each entry.
	\item Dataset conditions are enumerated into dictionaries by reaction roles, which we term categories, and ground-truth vectors are binned accordingly.\footnote{For consistency, individual reagents and conditions are referred to simply as labels, regardless of their identity. The terms label and bin are used interchangeably.} 
\end{enumerate}
With this procedure, significant trimming of the label spaces was achieved (see Table 1), while still maintaining deep and representative dictionaries. 
%
{\renewcommand{\arraystretch}{1.25}
\begin{table}[t!]
\caption{Summary of reaction sets studied.}
\vspace{0.1in}
\centering
\begin{tabular}{c c c c c}
    \hline
    name & reactions & raw labels & bins & categories \\
    \hline
    Suzuki & 145,413 & 3,315 & 118 & 5 \\
    C–N & 36,519 & 1,528 & 205 & 5 \\
    Negishi & 6,391 & 492 & 105 & 5 \\
    PKR & 2,749 & 335 & 83 & 8 \\
    \hline
\end{tabular}
\label{tab:rxn}
\end{table}}
\par \textbf{Learning task \& model setup}. Similarly to \citet{gao_using_2018}, we construct the learning problem as one of multi-label classification. Reactant and product graphs are fed as inputs to GNNs, which are trained to output binary condition vectors. The graphs are constructed using preprocessors from Chainer Chemistry (ChainerChem) \citep{tokui_chainer_2015}, which operate on RDKit mol objects \citep{open-source_rdkit_2006} calculated from dataset SMILES.
\par Our modeling studies test seven GNN architectures from the ChainerChem library. Each model contains two subnetworks that are jointly trained for the overall task. The first subnet is a graph processing network (GPN) that differs between architectures and forms the basis of their relative performances. The GPNs convert input graphs to learned molecular embeddings, which are concatenated to form the overall reaction vectors. These are passed as input to the second subnet, a multilayer perceptron (MLP), for the ultimate predictions. GPNs explored in this work include neural fingerprinting networks (NFPs) \citep{duvenaud_convolutional_2015}, gated graph sequence NNs (GGNNs) \citep{li_gated_2017}, message passing NNs (MPNNs) \citep{gilmer_neural_2017}, Weave module NNs (Weave) \citep{kearnes_molecular_2016}, relational graph attention networks (R-GATs) \citep{velickovic_graph_2018}, relational graph convolutional networks (R-GCNs) \citep{schlichtkrull_modeling_2017}, and renormalized spectral graph convolutional networks (RS-GCNs) \citep{kipf_semi-supervised_2017}.\footnote{Abbreviations used here are true to the original reports of each architecture; some differ from those in ChainerChem code.} Models are trained for 100 epochs using the Adam optimizer \citep{kingma_adam_2017}, sigmoid cross entropy loss, and an 80/10/10 train/validation/test split in all experiments. Further general modeling parameters and detailed hyperparameter settings for each model are included in the SM in Tables \ref{tab:computational} and \ref{tab:hyperparameters}.
%
{\renewcommand{\arraystretch}{1.1}
\begin{table*}[t]
\caption{Summary of top-1 ranking accuracies for all architectures across the four datasets.}
\vspace{0.1in}
\centering
\begin{tabular}{ c | c c c c c c c c c }
    \hline
    reaction & category & dummy & NFP & GGNN & MPNN & Weave & R-GAT & R-GCN & RS-GCN \\
    \hline
    \multirow{6}{3em}{\centering Suzuki} & \textbf{AER} & - & 0.1572 & 0.1297 & 0.0259 & 0.0388 & 0.0801 & \textbf{0.2767} & 0.0750 \\[0.9mm]
    & metal & 0.3777 & 0.5763 & 0.5291 & 0.4513 & 0.4759 & 0.4891 & \textbf{0.6306} & 0.4987 \\
    & ligand & 0.8722 & 0.8847 & 0.8811 & 0.8722 & 0.8724 & 0.8770 & \textbf{0.9036} & 0.8752 \\
    & base & 0.3361 & 0.4637 & 0.4377 & 0.3494 & 0.3640 & 0.4167 & \textbf{0.5455} & 0.4052 \\
    & solvent & 0.6377 & 0.6656 & 0.6656 & 0.6377 & 0.6381 & 0.6506 & \textbf{0.7049} & 0.6495 \\
    & additive & 0.9511 & 0.9560 & 0.9563 & 0.9507 & 0.9507 & 0.9524 & \textbf{0.9624} & 0.9521 \\[0.9mm]
    \hline
    \multirow{6}{3em}{\centering C–N} & \textbf{AER} & - & 0.2575 & 0.3178 & 0.0453 & 0.1048 & 0.1983 & \textbf{0.3453} & 0.1821 \\[0.9mm]
    & metal & 0.2452 & 0.5485 & 0.5847 & 0.3304 & 0.4261 & 0.5082 & \textbf{0.5989} & 0.4792 \\
    & ligand & 0.5219 & 0.6395 & 0.6789 & 0.5197 & 0.5327 & 0.6019 & \textbf{0.6981} & 0.5737 \\
    & base & 0.2479 & 0.5340 & 0.5710 & 0.3227 & 0.3909 & 0.4753 & \textbf{0.5932} & 0.4721 \\
    & solvent & 0.3219 & 0.4792 & 0.5348 & 0.3345 & 0.3690 & 0.4345 & \textbf{0.5647} & 0.4351 \\
    & additive & 0.8904 & 0.8934 & 0.8978 & 0.8904 & 0.8907 & 0.8912 & \textbf{0.8984} & 0.8934 \\[0.9mm]
    \hline
    \multirow{6}{3em}{\centering Negishi} & \textbf{AER} & - & 0.3071 & \textbf{0.4652} & 0.0916 & 0.0992 & 0.1539 & 0.4439 & 0.2228 \\[0.9mm]
    & metal & 0.2887 & 0.5470 & \textbf{0.6715} & 0.2887 & 0.3254 & 0.4067 & 0.6555 & 0.4833 \\
    & ligand & 0.7879 & 0.8485 & 0.8708 & 0.7879 & 0.7879 & 0.7974 & \textbf{0.8724} & 0.8102 \\
    & temperature & 0.3317 & 0.4864 & \textbf{0.6459} & 0.3732 & 0.4163 & 0.4035 & 0.6188 & 0.4864 \\
    & solvent & 0.6938 & 0.8596 & 0.8852 & 0.8150 & 0.7911 & 0.8262 & \textbf{0.8868} & 0.8278 \\
    & additive & 0.8309 & 0.8501 & \textbf{0.8820} & 0.8309 & 0.8309 & 0.8341 & 0.8724 & 0.8421 \\[0.9mm]
    \hline
    \multirow{9}{3em}{\centering PKR} & \textbf{AER} & - & 0.2400 & \textbf{0.4377} & -0.0294 & 0.1209 & 0.0825 & 0.3973 & 0.2265 \\[0.9mm]
    & metal & 0.4302 & 0.6340 & 0.7094 & 0.4302 & 0.4943 & 0.4566 & \textbf{0.7132} & 0.5774 \\
    & ligand & 0.8792 & 0.8981 & \textbf{0.9094} & 0.8792 & 0.8868 & 0.8792 & 0.9057 & 0.9019 \\
    & temperature & 0.2830 & 0.4415 & \textbf{0.6642} & 0.3358 & 0.4000 & 0.3283 & 0.6528 & 0.4755 \\
    & solvent & 0.3321 & 0.5358 & \textbf{0.7396} & 0.3887 & 0.3774 & 0.4000 & 0.6792 & 0.5472 \\
    & activator & 0.6906 & 0.7774 & \textbf{0.8679} & 0.6906 & 0.7094 & 0.6755 & 0.8415 & 0.7660 \\
    & CO (g) & 0.7245 & 0.7849 & 0.8642 & 0.4755 & 0.6906 & 0.7208 & \textbf{0.8717} & 0.7434 \\
    & additive & 0.9057 & 0.8943 & 0.8981 & 0.9057 & \textbf{0.9132} & 0.9057 & 0.8906 & 0.8981 \\
    & pressure & 0.6528 & 0.8264 & \textbf{0.8679} & 0.8302 & 0.8415 & 0.8302 & 0.8491 & 0.8415 \\[0.9mm]
    \hline
\end{tabular}
\label{tab:results_top-1}
\end{table*}
}
\par \textbf{Model output and evaluation}. We analyze the success of our models in terms of their accuracy in predicting the ground truth label for each reaction role. In practice, the outputs are simply probability vectors corresponding to the full reaction dictionaries. These are postprocessed by sorting into categorical sub-dictionaries, and the final output is a list of labels for each category, ranked by their probability scores. A category’s prediction is classified as accurate if the ground truth label is identified in the model’s top-$k$ predicted rankings. Here, we consider top-1 and top-3 predictions, though this is amenable to preference. Categorical accuracy ($A_c$) is defined as follows:
\begin{equation}
    A_c = \frac{1}{N} \sum_{i=1}^{N} \mathbb{1} [P_i \cap Y_i]
    \label{eqn:acc}
\end{equation}
where $P_i$ and $Y_i$ are the sets of predicted and ground truth labels of the $i$-th sample, respectively, and $N$ is the number of samples in the test set \citep{wu_unified_2017}. 
\par We directly compare model performances to a dummy predictor (dummy) that always suggests the most frequently occurring label(s) from each category of a dataset. Since there is variable class-imbalance between categories \citep{cui_class-balanced_2019} (see SM for full distributions), instead of averaging $A_c$ values for a reaction model we calculate their average error reduction (AER) from baseline. We use AER to compare overall architecture performances on each task, and simply define it as follows:
\begin{equation}
    \textrm{AER} = \frac{1}{C} \sum_{c=1}^{C} \frac{A^{g}_c - A^{d}_c}{1 - A^{d}_c}
    \label{eqn:aer}
\end{equation}
where $A^{g}_c$ and $A^{d}_c$ are the accuracies of the graph network and dummy model in the $c$-th category, respectively, and $C$ is the number of categories in the dataset dictionary.
\section{Results and Analysis}
\label{sec:results}
The top-1 ranking accuracy of each architecture on all four tasks is presented in Table \ref{tab:results_top-1}. An expanded results table with top-3 performances is included in the SM in Table \ref{tab:results_top-3}. Several of the tested networks provide strong general accuracy and significant AERs over baseline, with GGNNs and R-GCNs performing best in most cases. Categorical trends can be noted for each reaction, summarized below:
\begin{enumerate}
    \item For the Suzuki dataset, when compared to baseline our models best improve metal and base predictions, but struggle with ligand and solvent.
    \item For C–N couplings, additives prove challenging, while good improvements are made otherwise.
    \item For Negishi couplings, models perform very well with metal, temperature, and solvent predictions, but again struggle with additives.
    \item For the PKR dataset, strong improvements are made with temperature, solvent, activator, and pressure, while only minor gains are seen for ligand and additive.
\end{enumerate}
It is interesting to note that certain architectures behave differently between reactions, perhaps owing to model size, dataset size, and/or the chemical space within them. Though not included here, future studies will investigate the effects of specific convolution types in each architecture.
\par
\begin{figure}[t]
    \centering
    \caption{R-GCN activation visualization and predictions for a selected random reaction from the C–N coupling test set. Darker highlights indicate higher atom activation.}
    \vspace{0.1in}
    \includegraphics[width = 0.45\textwidth]{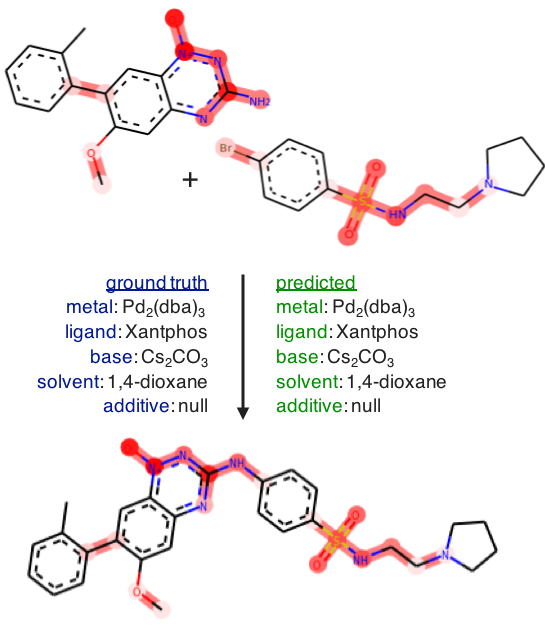}
    \label{fig:viz}
\end{figure}

 \textbf{Model interpretation}. To gain insight into the chemical information being learned in our modeling, we investigated the graph features leading to the observed predictions. To visualize groups of atoms most ``informative" to the model readout, we extracted atom feature vectors from R-GCNs, the top performer from our modeling studies. An example visualization of a C–N coupling is shown in Figure \ref{fig:viz}. In line with chemical intuition, the strongest activation comes from heteroatom (non-carbon) groups surrounding the reaction sites in the reactants and product. Additional activation is seen in distal groups that one might expect to interfere with the desired reaction. In this example, all five category labels are predicted correctly.
\section{Discussion \& Outlook}
\label{sec:discussion}

\textbf{Advantages of the approach}. As noted in Sections \ref{sec:intro} and \ref{sec:approach_related}, the approach presented here has several major benefits:
\begin{enumerate}
    \item Reaction-specific modeling offers fine-grained learning and circumvents the sparsity of out-of-scope reactions. 
    \item Expert-level label categorization ensures chemically reasonable outputs and reduces noise, a documented limitation of prior methods that we improve here.
    \item Model readouts can be visualized, increasing the interpretability of molecular deep learning.
\end{enumerate}
\par \textbf{Limitations}. We also acknowledge certain limitations of the method. Since predictions are strictly structure-based, there is an inherent limit on the achievable accuracy. Several other features are expected to be informative for modeling what is really historical reaction data. Preliminary experiments have shown that metadata such as publication year does slightly increase model accuracy, but we exclude these features here since they have no physical bearing on reactivity. Further, we do not consider ``close match" predictions. This results in more rigorous accuracy metrics, but discourages potentially useful conditions similar to the ground truth from being suggested. Altogether, we see several opportunities for improvement to be explored in future works.
\par \textbf{Final remarks}. In summary, we present a novel approach using GNNs to predict organic chemical reaction conditions. Categorizing labels by reaction role, we achieve top-1 ranking accuracies of up to 96\% and reduce baseline error by as much as 47\%. We find the approach general across four valuable reaction types, with GGNNs and R-GCNs performing well overall. Trained models can be readily applied to suggest context-specific experimental conditions, representing a significant contribution to synthetic chemistry. Ongoing work is focused on optimizing GNN architectures, adding to the available reaction types, and integrating label correlations in modeling. We expect the tools reported here to be of great value in molecular machine learning, including for computer-aided synthesis planning and drug development.
\section*{Acknowledgements}
We thank the reviewers for their insightful comments and Prof Pietro Perona for mentorship guidance and helpful discussions on this work. Fellowship support was provided by the NSF (M.R.M., T.J.D. Grant No. DGE- 1144469). S.E.R. is a Heritage Medical Research Investigator. Financial support from the Research Corporation Cottrell Scholars Program is acknowledged.

\bibliography{manuscript}

\begin{thebibliography}{30}
\providecommand{\natexlab}[1]{#1}
\providecommand{\url}[1]{\texttt{#1}}
\expandafter\ifx\csname urlstyle\endcsname\relax
  \providecommand{\doi}[1]{doi: #1}\else
  \providecommand{\doi}{doi: \begingroup \urlstyle{rm}\Url}\fi

\bibitem[Bariwal \& Van~der Eycken(2013)Bariwal and Van~der
  Eycken]{bariwal_cn_2013}
Bariwal, J. and Van~der Eycken, E.
\newblock C–{N} bond forming cross-coupling reactions: an overview.
\newblock \emph{Chemical Society Reviews}, 42\penalty0 (24):\penalty0 9283,
  2013.
\newblock ISSN 0306-0012, 1460-4744.
\newblock \doi{10.1039/c3cs60228a}.
\newblock URL \url{http://xlink.rsc.org/?DOI=c3cs60228a}.

\bibitem[Blaschke et~al.(2018)Blaschke, Olivecrona, Engkvist, Bajorath, and
  Chen]{blaschke_application_2018}
Blaschke, T., Olivecrona, M., Engkvist, O., Bajorath, J., and Chen, H.
\newblock Application of {Generative} {Autoencoder} in {De} {Novo} {Molecular}
  {Design}.
\newblock \emph{Molecular Informatics}, 37\penalty0 (1-2):\penalty0 1700123,
  2018.
\newblock ISSN 1868-1751.
\newblock \doi{10.1002/minf.201700123}.
\newblock URL
  \url{https://onlinelibrary.wiley.com/doi/abs/10.1002/minf.201700123}.

\bibitem[Coley et~al.(2017)Coley, Barzilay, Jaakkola, Green, and
  Jensen]{coley_prediction_2017}
Coley, C.~W., Barzilay, R., Jaakkola, T.~S., Green, W.~H., and Jensen, K.~F.
\newblock Prediction of {Organic} {Reaction} {Outcomes} {Using} {Machine}
  {Learning}.
\newblock \emph{ACS Central Science}, 3\penalty0 (5):\penalty0 434--443, May
  2017.
\newblock ISSN 2374-7943.
\newblock \doi{10.1021/acscentsci.7b00064}.
\newblock URL \url{https://doi.org/10.1021/acscentsci.7b00064}.

\bibitem[Coley et~al.(2018)Coley, Green, and Jensen]{coley_machine_2018}
Coley, C.~W., Green, W.~H., and Jensen, K.~F.
\newblock Machine {Learning} in {Computer}-{Aided} {Synthesis} {Planning}.
\newblock \emph{Accounts of Chemical Research}, 51\penalty0 (5):\penalty0
  1281--1289, May 2018.
\newblock ISSN 0001-4842.
\newblock \doi{10.1021/acs.accounts.8b00087}.
\newblock URL \url{https://doi.org/10.1021/acs.accounts.8b00087}.

\bibitem[Coley et~al.(2019)Coley, Jin, Rogers, Jamison, Jaakkola, Green,
  Barzilay, and Jensen]{coley_graph-convolutional_2019}
Coley, C.~W., Jin, W., Rogers, L., Jamison, T.~F., Jaakkola, T.~S., Green,
  W.~H., Barzilay, R., and Jensen, K.~F.
\newblock A graph-convolutional neural network model for the prediction of
  chemical reactivity.
\newblock \emph{Chemical Science}, 10\penalty0 (2):\penalty0 370--377, January
  2019.
\newblock ISSN 2041-6539.
\newblock \doi{10.1039/C8SC04228D}.
\newblock URL
  \url{https://pubs.rsc.org/en/content/articlelanding/2019/sc/c8sc04228d}.

\bibitem[Cui et~al.(2019)Cui, Jia, Lin, Song, and
  Belongie]{cui_class-balanced_2019}
Cui, Y., Jia, M., Lin, T.-Y., Song, Y., and Belongie, S.
\newblock Class-{Balanced} {Loss} {Based} on {Effective} {Number} of {Samples}.
\newblock \emph{arXiv:1901.05555 [cs]}, January 2019.
\newblock URL \url{http://arxiv.org/abs/1901.05555}.
\newblock arXiv: 1901.05555.

\bibitem[Duvenaud et~al.(2015)Duvenaud, Maclaurin, Aguilera-Iparraguirre,
  Gómez-Bombarelli, Hirzel, Aspuru-Guzik, and
  Adams]{duvenaud_convolutional_2015}
Duvenaud, D., Maclaurin, D., Aguilera-Iparraguirre, J., Gómez-Bombarelli, R.,
  Hirzel, T., Aspuru-Guzik, A., and Adams, R.~P.
\newblock Convolutional {Networks} on {Graphs} for {Learning} {Molecular}
  {Fingerprints}.
\newblock \emph{arXiv:1509.09292 [cs, stat]}, November 2015.
\newblock URL \url{http://arxiv.org/abs/1509.09292}.
\newblock arXiv: 1509.09292.

\bibitem[Elton et~al.(2019)Elton, Boukouvalas, Fuge, and
  Chung]{elton_deep_2019}
Elton, D.~C., Boukouvalas, Z., Fuge, M.~D., and Chung, P.~W.
\newblock Deep learning for molecular design—a review of the state of the
  art.
\newblock \emph{Molecular Systems Design \& Engineering}, 4\penalty0
  (4):\penalty0 828--849, 2019.
\newblock ISSN 2058-9689.
\newblock \doi{10.1039/C9ME00039A}.
\newblock URL \url{http://xlink.rsc.org/?DOI=C9ME00039A}.

\bibitem[Gao et~al.(2018)Gao, Struble, Coley, Wang, Green, and
  Jensen]{gao_using_2018}
Gao, H., Struble, T.~J., Coley, C.~W., Wang, Y., Green, W.~H., and Jensen,
  K.~F.
\newblock Using {Machine} {Learning} {To} {Predict} {Suitable} {Conditions} for
  {Organic} {Reactions}.
\newblock \emph{ACS Central Science}, 4\penalty0 (11):\penalty0 1465--1476,
  November 2018.
\newblock ISSN 2374-7943.
\newblock \doi{10.1021/acscentsci.8b00357}.
\newblock URL \url{https://doi.org/10.1021/acscentsci.8b00357}.

\bibitem[Gilmer et~al.(2017)Gilmer, Schoenholz, Riley, Vinyals, and
  Dahl]{gilmer_neural_2017}
Gilmer, J., Schoenholz, S.~S., Riley, P.~F., Vinyals, O., and Dahl, G.~E.
\newblock Neural {Message} {Passing} for {Quantum} {Chemistry}.
\newblock \emph{arXiv:1704.01212 [cs]}, June 2017.
\newblock URL \url{http://arxiv.org/abs/1704.01212}.
\newblock arXiv: 1704.01212.

\bibitem[Kearnes et~al.(2016)Kearnes, McCloskey, Berndl, Pande, and
  Riley]{kearnes_molecular_2016}
Kearnes, S., McCloskey, K., Berndl, M., Pande, V., and Riley, P.
\newblock Molecular {Graph} {Convolutions}: {Moving} {Beyond} {Fingerprints}.
\newblock \emph{Journal of Computer-Aided Molecular Design}, 30\penalty0
  (8):\penalty0 595--608, August 2016.
\newblock ISSN 0920-654X, 1573-4951.
\newblock \doi{10.1007/s10822-016-9938-8}.
\newblock URL \url{http://arxiv.org/abs/1603.00856}.
\newblock arXiv: 1603.00856.

\bibitem[Khand et~al.(1973)Khand, Knox, Pauson, Watts, and
  Foreman]{khand_organocobalt_1973}
Khand, I.~U., Knox, G.~R., Pauson, P.~L., Watts, W.~E., and Foreman, M.~I.
\newblock Organocobalt complexes. {Part} {II}. {Reaction} of
  acetylenehexacarbonyldicobalt complexes, ({R1C2R2}){Co2}({CO})6, with
  norbornene and its derivatives.
\newblock \emph{Journal of the Chemical Society, Perkin Transactions 1},
  0\penalty0 (0):\penalty0 977--981, January 1973.
\newblock ISSN 1364-5463.
\newblock \doi{10.1039/P19730000977}.
\newblock URL
  \url{https://pubs.rsc.org/en/content/articlelanding/1973/p1/p19730000977}.

\bibitem[Kingma \& Ba(2017)Kingma and Ba]{kingma_adam_2017}
Kingma, D.~P. and Ba, J.
\newblock Adam: {A} {Method} for {Stochastic} {Optimization}.
\newblock \emph{arXiv:1412.6980 [cs]}, January 2017.
\newblock URL \url{http://arxiv.org/abs/1412.6980}.
\newblock arXiv: 1412.6980.

\bibitem[Kipf \& Welling(2017)Kipf and Welling]{kipf_semi-supervised_2017}
Kipf, T.~N. and Welling, M.
\newblock Semi-{Supervised} {Classification} with {Graph} {Convolutional}
  {Networks}.
\newblock \emph{arXiv:1609.02907 [cs, stat]}, February 2017.
\newblock URL \url{http://arxiv.org/abs/1609.02907}.
\newblock arXiv: 1609.02907.

\bibitem[Li et~al.(2017)Li, Tarlow, Brockschmidt, and Zemel]{li_gated_2017}
Li, Y., Tarlow, D., Brockschmidt, M., and Zemel, R.
\newblock Gated {Graph} {Sequence} {Neural} {Networks}.
\newblock \emph{arXiv:1511.05493 [cs, stat]}, September 2017.
\newblock URL \url{http://arxiv.org/abs/1511.05493}.
\newblock arXiv: 1511.05493.

\bibitem[Mahatthananchai et~al.(2012)Mahatthananchai, Dumas, and
  Bode]{mahatthananchai_catalytic_2012}
Mahatthananchai, J., Dumas, A.~M., and Bode, J.~W.
\newblock Catalytic {Selective} {Synthesis}.
\newblock \emph{Angewandte Chemie International Edition}, 51\penalty0
  (44):\penalty0 10954--10990, October 2012.
\newblock ISSN 14337851.
\newblock \doi{10.1002/anie.201201787}.
\newblock URL \url{http://doi.wiley.com/10.1002/anie.201201787}.

\bibitem[Mater \& Coote(2019)Mater and Coote]{mater_deep_2019}
Mater, A.~C. and Coote, M.~L.
\newblock Deep {Learning} in {Chemistry}.
\newblock \emph{Journal of Chemical Information and Modeling}, 59\penalty0
  (6):\penalty0 2545--2559, June 2019.
\newblock ISSN 1549-9596, 1549-960X.
\newblock \doi{10.1021/acs.jcim.9b00266}.
\newblock URL \url{https://pubs.acs.org/doi/10.1021/acs.jcim.9b00266}.

\bibitem[Miyaura \& Suzuki(1995)Miyaura and
  Suzuki]{miyaura_palladium-catalyzed_1995}
Miyaura, N. and Suzuki, A.
\newblock Palladium-{Catalyzed} {Cross}-{Coupling} {Reactions} of {Organoboron}
  {Compounds}.
\newblock \emph{Chemical Reviews}, 95\penalty0 (7):\penalty0 2457--2483,
  November 1995.
\newblock ISSN 0009-2665, 1520-6890.
\newblock \doi{10.1021/cr00039a007}.
\newblock URL \url{https://pubs.acs.org/doi/abs/10.1021/cr00039a007}.

\bibitem[Negishi et~al.(1977)Negishi, King, and
  Okukado]{negishi_selective_1977}
Negishi, E., King, A.~O., and Okukado, N.
\newblock Selective carbon-carbon bond formation via transition metal
  catalysis. 3. {A} highly selective synthesis of unsymmetrical biaryls and
  diarylmethanes by the nickel- or palladium-catalyzed reaction of aryl- and
  benzylzinc derivatives with aryl halides.
\newblock \emph{The Journal of Organic Chemistry}, 42\penalty0 (10):\penalty0
  1821--1823, May 1977.
\newblock ISSN 0022-3263, 1520-6904.
\newblock \doi{10.1021/jo00430a041}.
\newblock URL \url{https://pubs.acs.org/doi/abs/10.1021/jo00430a041}.

\bibitem[Open-Source(2006)]{open-source_rdkit_2006}
Open-Source.
\newblock {RDKit}: {Open}-{Source} {Cheminformatics} {Software}, 2006.

\bibitem[Reaxys()]{reaxys_reaxys_nodate}
Reaxys.
\newblock Reaxys.
\newblock URL \url{https://new.reaxys.com/}.

\bibitem[Schlichtkrull et~al.(2017)Schlichtkrull, Kipf, Bloem, Berg, Titov, and
  Welling]{schlichtkrull_modeling_2017}
Schlichtkrull, M., Kipf, T.~N., Bloem, P., Berg, R. v.~d., Titov, I., and
  Welling, M.
\newblock Modeling {Relational} {Data} with {Graph} {Convolutional} {Networks}.
\newblock \emph{arXiv:1703.06103 [cs, stat]}, October 2017.
\newblock URL \url{http://arxiv.org/abs/1703.06103}.
\newblock arXiv: 1703.06103.

\bibitem[Schwaller et~al.(2019)Schwaller, Probst, Vaucher, Nair, Laino, and
  Reymond]{schwaller_data-driven_2019}
Schwaller, P., Probst, D., Vaucher, A.~C., Nair, V.~H., Laino, T., and Reymond,
  J.-L.
\newblock Data-{Driven} {Chemical} {Reaction} {Classification},
  {Fingerprinting} and {Clustering} using {Attention}-{Based} {Neural}
  {Networks}.
\newblock December 2019.
\newblock \doi{10.26434/chemrxiv.9897365.v2}.
\newblock URL
  \url{https://chemrxiv.org/articles/Data-Driven_Chemical_Reaction_Classification_with_Attention-Based_Neural_Networks/9897365}.

\bibitem[Segler et~al.(2018)Segler, Preuss, and Waller]{segler_planning_2018}
Segler, M. H.~S., Preuss, M., and Waller, M.~P.
\newblock Planning chemical syntheses with deep neural networks and symbolic
  {AI}.
\newblock \emph{Nature}, 555\penalty0 (7698):\penalty0 604--610, March 2018.
\newblock ISSN 0028-0836, 1476-4687.
\newblock \doi{10.1038/nature25978}.
\newblock URL \url{http://www.nature.com/doifinder/10.1038/nature25978}.

\bibitem[Skoraczyński et~al.(2017)Skoraczyński, Dittwald, Miasojedow,
  Szymkuć, Gajewska, Grzybowski, and Gambin]{skoraczynski_predicting_2017}
Skoraczyński, G., Dittwald, P., Miasojedow, B., Szymkuć, S., Gajewska, E.~P.,
  Grzybowski, B.~A., and Gambin, A.
\newblock Predicting the outcomes of organic reactions via machine learning:
  are current descriptors sufficient?
\newblock \emph{Scientific Reports}, 7, June 2017.
\newblock ISSN 2045-2322.
\newblock \doi{10.1038/s41598-017-02303-0}.
\newblock URL \url{https://www.ncbi.nlm.nih.gov/pmc/articles/PMC5472585/}.

\bibitem[Tokui et~al.(2015)Tokui, Oono, Hido, and Clayton]{tokui_chainer_2015}
Tokui, S., Oono, K., Hido, S., and Clayton, J.
\newblock Chainer: a {Next}-{Generation} {Open} {Source} {Framework} for {Deep}
  {Learning}.
\newblock 2015.
\newblock URL
  \url{https://chainer-chemistry.readthedocs.io/en/latest/index.html}.

\bibitem[Veličković et~al.(2018)Veličković, Cucurull, Casanova, Romero,
  Liò, and Bengio]{velickovic_graph_2018}
Veličković, P., Cucurull, G., Casanova, A., Romero, A., Liò, P., and Bengio,
  Y.
\newblock Graph {Attention} {Networks}.
\newblock \emph{arXiv:1710.10903 [cs, stat]}, February 2018.
\newblock URL \url{http://arxiv.org/abs/1710.10903}.
\newblock arXiv: 1710.10903.

\bibitem[Weininger(1988)]{weininger_smiles_1988}
Weininger, D.
\newblock {SMILES}, a chemical language and information system. 1.
  {Introduction} to methodology and encoding rules.
\newblock \emph{Journal of Chemical Information and Modeling}, 28\penalty0
  (1):\penalty0 31--36, February 1988.
\newblock ISSN 1549-9596.
\newblock \doi{10.1021/ci00057a005}.
\newblock URL \url{https://pubs.acs.org/doi/abs/10.1021/ci00057a005}.

\bibitem[Withnall et~al.(2020)Withnall, Lindelöf, Engkvist, and
  Chen]{withnall_building_2020}
Withnall, M., Lindelöf, E., Engkvist, O., and Chen, H.
\newblock Building attention and edge message passing neural networks for
  bioactivity and physical–chemical property prediction.
\newblock \emph{Journal of Cheminformatics}, 12\penalty0 (1), December 2020.
\newblock ISSN 1758-2946.
\newblock \doi{10.1186/s13321-019-0407-y}.
\newblock URL
  \url{https://jcheminf.biomedcentral.com/articles/10.1186/s13321-019-0407-y}.

\bibitem[Wu \& Zhou(2017)Wu and Zhou]{wu_unified_2017}
Wu, X.-Z. and Zhou, Z.-H.
\newblock A {Unified} {View} of {Multi}-{Label} {Performance} {Measures}.
\newblock \emph{arXiv:1609.00288 [cs]}, September 2017.
\newblock URL \url{http://arxiv.org/abs/1609.00288}.
\newblock arXiv: 1609.00288.

\end{thebibliography}
\bibliographystyle{icml2020}

\newpage
\onecolumn
\icmltitle{Supplementary Material: Graph Neural Networks for the Prediction of Organic Reaction Conditions}



\icmlsetsymbol{equal}{*}

\begin{icmlauthorlist}
\icmlauthor{Serim Ryou}{equal,cvl}
\icmlauthor{Michael R. Maser}{equal,cce}
\icmlauthor{Alexander Y. Cui}{equal,cms}
\icmlauthor{Travis J. DeLano}{cce}
\icmlauthor{Yisong Yue}{cms}
\icmlauthor{Sarah E. Reisman}{cce}
\end{icmlauthorlist}

\icmlaffiliation{cce}{Division of Chemistry and Chemical Engineering, California Institute of Technology, Pasadena, California, USA}
\icmlaffiliation{cvl}{Computational Vision Lab, California Institute of Technology, Pasadena, California, USA}
\icmlaffiliation{cms}{Department of Computing and Mathematical Sciences, California Institute of Technology, Pasadena, California, USA}

\icmlcorrespondingauthor{Michael R. Maser}{mmaser@caltech.edu}
\icmlcorrespondingauthor{Sarah E. Reisman}{reisman@caltech.edu}

\vskip 0.3in




\beginsupplement

\section{Data preparation}
\label{sec:data}

All datasets used herein were obtained from queries to the Reaxys® database \cite{reaxys_reaxys_nodate}. Results from both journals and patents are included for all reaction types. An expanded Table 1 with depictions of the four reaction queries is shown in Table \ref{tab:queries}.  

{\renewcommand{\arraystretch}{1.75}
\begin{table}[H]
\caption{Full summary of reaction sets studied with Reaxys® query depictions.}
\vspace{0.1in}
\centering
\begin{tabular}{c c c c c c}
    \hline
    name & query & reactions & bins & reactions/bin & categories \\
    \hline
    Suzuki & \includegraphics[scale=0.7]{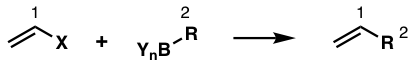} & 145,413 & 118 & 1232.3 & 5 \\
    C–N & \includegraphics[scale=0.7]{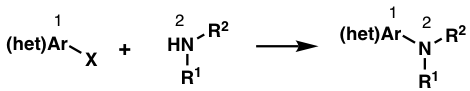} & 36,519 & 205 & 178.1 & 5 \\
    Negishi & \includegraphics[scale=0.7]{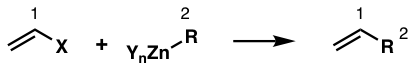} & 6,391 & 105 & 60.9 & 5 \\
    PKR & \includegraphics[scale=0.7]{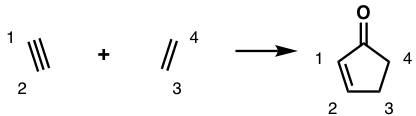} & 2,749 & 83 & 33.1 & 8 \\
    \hline
\end{tabular}
\label{tab:queries}
\end{table}}

For each reaction, the atoms involved in cross-coupling bond formation are enumerated in reactants and products, specified in the Reaxys® queries \emph{via} atom mapping. The number of reactions in Table \ref{tab:queries} refers to counts after pre-processing. General pre-processing details used for all reactions are included below, followed by reaction-specific procedures. It should be noted that the four reactions modeled here share a similar format, typically containing one or two ``coupling partners'' (i.e. reactants) that react to form a single product upon treatment with a set of categorized conditions. In principle, the modeling framework described herein can be applied to any reaction format, provided a condition dictionary is written. Studies applying the current framework to reactions of other formats are currently ongoing.
\par \textbf{General pre-processing}. In all datasets, reactions without reactant or product structures, condition information, or reaction yields are removed unless otherwise specified. Duplicate condition information is removed such that each reaction contained only unique entries for each dictionary category (\emph{vide infra}). It is possible that this duplicate information is meant to signify multiple equivalents of reaction components (i.e. stoichiometry), but given the unstructured nature of this feature in Reaxys®, we do not consider it here. After trimming, condition entries are pooled and all unique values extracted with their frequency of appearance in the trimmed dataset to construct the full length reaction dictionary. The dictionary is truncated at a cumulative 95\% coverage of instance frequencies. This serves to avoid both sparsity in label bins and unreasonably lengthy curation. All resulting entries are assigned an identifier for their reaction role, which we term categories. The dictionaries are then sorted into categories for further processing. Reagents that could serve multiple reaction roles are assigned each plausible identifier and copied into each respective category. 
\par Once categorized, the frequencies of each bin are recalculated within the category, and each category is again truncated at 95\% total coverage. At the data scale studied here, reagent bins outside of this threshold typically appear in the dataset under 10 times total. We therefore exclude them to avoid unnecessary dimensionality in the label space and reduce overfitting. Even still, class-imbalance within the categories is pronounced \citep{cui_class-balanced_2019}, as evidenced by the ``long-tail'' distributions provided in Section \ref{sec:EDA}. Categories are combed to identify any bins with unique names that refer to the same reagent, often by misspelling, abbreviation, etc. At this stage, all unique species are assigned a bin label within their category, and the categories are recombined. This constitutes the final reaction dictionary and defines the label space for our prediction task. Appropriate categories were also assigned ``null'' bins to indicate reactions not specifying labels in that category. 
\par \textbf{Additional Suzuki pre-processing}. The initial export contained 154,634 data points at time of download. Reactions without yields (\texttildelow1,500) or solvent (\texttildelow2,700) were removed. Reactions with more or less than 2 reactants, and more than one product (\texttildelow6,800) were removed. The final dataset contained 145,413 reactions with a dictionary of 118 bins in the categories metal (M), ligand (L), base (B), solvent (S), and additive (A). Roughly 25\% of reactions did not have temperatures specified, so this feature was dropped.
\par \textbf{Additional C–N pre-processing}. The initial export contained 39,902 data points at time of download. Reactions without solvent were retained, as this dataset included non-catalytic C–N couplings such as S$_N$Ar reactions, often run in the neat amine reactant. Reactions with more or less than 2 reactants and 1 product (\texttildelow3,000) were removed. Reactions with more than 4 reagents (\texttildelow250) were removed. The final dataset contained 36,519 reactions with a dictionary of 205 bins in the categories metal (M), ligand (L), base (B), solvent (S), and additive (A). Roughly 30\% of reactions did not have temperatures specified, so this feature was dropped.
\par \textbf{Additional Negishi pre-processing}. The initial export contained 11,388 data points at time of download. Reactions without yields (\texttildelow3,500) or solvent (\texttildelow350) were removed. Reactions with more than 2 reactants, 3 solvents, and/or 4 reagents (\texttildelow1,200) were removed. The final dataset contained 6,391 reactions with a dictionary of 105 bins in the categories metal (M), ligand (L), temperature (T), solvent (S), and additive (A). Almost 90\% of reactions had temperatures specified, so this feature was retained. Those with unspecified temperatures were assumed to occur ambiently and assigned as 20 °C.
\par \textbf{Additional PKR pre-processing}. The initial export contained 4,275 data points at time of download. Both inter- and intramolecular reactions were retained, so reactions contained either 1 or 2 reactants. Reactions without yields (\texttildelow1,000) were removed. Reactions with more than 2 reactants, 3 solvents, and/or 4 reagents (\texttildelow500) were removed. The final dataset contained 2,749 reactions with a dictionary of 83 bins in the categories metal (M), ligand (L), temperature (T), solvent (S), additive (O), activator (A), gas (G), and pressure (P). The gas category is a binary identifier for the use of a carbon monoxide (CO) atmosphere.
\section{Computational details and hyperparameters}
\label{sec:computational}
\par All neural network (NN) architectures tested herein were used directly from the Chainer Chemistry (ChainerChem) library \citep{tokui_chainer_2015}, modified only as needed to fit each dataset task. In all cases, a graph processing network (GPN) was selected and combined with a dense multi-layer perceptron (MLP), which were trained together as a joint network. All models were trained for 100 epochs on 1 NVIDIA K80 GPU device, unless otherwise specified. Training and test sets were held consistent between models for each reaction dataset. This was done by first splitting each dataset into 90/10 train/test, then splitting the training set into 90/10 train/validation, resulting in a final split of 81/9/10 train/validation/test overall. A dummy predictor that always predicts the most frequent bin in each label category was also created for each dataset as a baseline performance reference. 
\par General parameters and default hyperparameter settings are summarized in Table \ref{tab:computational}, which are held constant across all models and datasets unless otherwise specified in Table \ref{tab:hyperparameters}. Every attempt was made to keep shared hyperparameters consistent between model types, and the majority were set to defaults. However, there were certain cases where this resulted in excessive memory requirements and crashes during model training. To adapt to these cases, parameters such as the hidden dimension (hidden{\_}dim) were incrementally decreased until training was successful. These cases are noted in Table \ref{tab:hyperparameters}.
\par Model names listed follow those from original references (see main text), names in parentheses refer to those used by Chainer functions. It should be noted that while the predicted vectors contain a single label from each category, it is possible that the ground truth contains more than one or zero. We add a null label to each sub-dictionary to handle the zero case, and with multiple ground truths we treat a category’s prediction as accurate if any are correctly identified. The null case was found commonly in ligand and additive categories, where a null ligand often resulted from use of a pre-ligated metal source. The multi-output scenario arose most frequently in the form of mixed solvent systems.
\begin{table}[H]
\caption{Computational details and general parameters used for all models.}
\vspace{0.1in}
\centering
\begin{tabular}{ c | c | c }
    \hline
    parameter & value & description \\
    \hline 
    loss & sigmoid cross entropy & loss function used for training \\
    optimizer & Adam & model optimization algorithm \\
    train/valid/test & 81/9/10 & data splitting \\
    batch size & 32 & batch size used for gradient calculations \\
    epochs & 100 & number of training epochs \\
    out{\_}dim & 128 & number of units in the readout \\
    hidden{\_}dim & 128 & number of units in the hidden layers \\
    n{\_}layers & 4 & number of convolutional layers \\
    n{\_}atom{\_}types & 117 & number of allowed atom types \\
    concat{\_}hidden & False & readouts concatenated at each layer \\
    \hline
\end{tabular}
\label{tab:computational}
\end{table}
\begin{table}[H]
\caption{Additional model-specific hyperparameter settings.}
\vspace{0.1in}
\centering
\begin{tabular}{ c | c c c }
    \hline
    \multirow{1}{5em}{\centering model} & hyperparameter & value & description\\
    \hline
    \multirow{1}{5em}{\centering NFP} & max{\_}degree & 6 & max degree of atom nodes in the graph \\
    \hline
    \multirow{3}{5em}{\centering GGNN} & 
    weight{\_}tying & True &  use weight tying \\
    & num{\_}edge{\_}type & 4 & \multirow{2}{20em}{\centering edge (i.e. bond) types allowed (4 includes single, double, triple and aromatic)} \\
    \\
    \hline
    \multirow{6}{5em}{\centering MPNN} & weight{\_}tying & True &  use weight tying \\ 
    & message{\_}func & `edgenet' & message function \\
    & readout{\_}func & `set2set' & readout function \\
    & hidden{\_}dim & 16 & default 128 required excessive memory \\
    & batch size & 8 & default 32 required excessive memory \\
    & epochs & 32 \& 5 & \multirow{2}{20em}{\centering Suzuki \& C–N; memory errors found at higher epochs; validation loss had converged} \\
    \\
    \hline
    \multirow{6}{5em}{\centering Weave (WeaveNet)} & weave{\_}channels & 200 & weave channel dimensionality \\
    & n{\_}atom & 20 & number of atoms in input arrays \\
    & n{\_}sub{\_}layer & 1 & number of layers in each pairing layer \\
    & readout{\_}mode & `sum' & readout mode \\
    & epochs & 10 & \multirow{2}{20em}{\centering Suzuki only; memory errors found at higher epochs; validation loss had converged} \\
    \\
    \hline
    \multirow{6}{5em}{\centering R-GAT (RelGAT)} & n{\_}heads & 3 & number of multi-head attentions \\
    & negative{\_}slope & 0.2 & LeakyRELU negative angle \\
    & dropout{\_}ratio & -1. & dropout for normalized attention coefficients \\
    & softmax{\_}mode & across & method for taking softmax over logits \\
    & concat{\_}heads & False & concatenate multi-head attentions \\
    & weight{\_}tying & False & use weight tying \\
    & hidden{\_}dim & 12 & default 128 required excessive memory \\
    \hline
    \multirow{4}{5em}{\centering R-GCN (RelGCN)} & out{\_}channels & 128 & output feature vector dimensionality \\
    & ch{\_}list & None & channels in update layers \\
    & input{\_}type & `int' & input vector type \\
    & scale{\_}adj & True & normalize adjacency matrix \\
    \hline
    \multirow{3}{5em}{\centering RS-GCN (RSGCN)} & use{\_}batch{\_}norm & False & apply batch normalization after convolutions \\
    & readout & None & readout mode (None defaults to `sum') \\
    & dropout{\_}ratio & 0.5 & dropout function ratio \\
    \hline
    \multirow{4}{5em}{\centering MLP} & out{\_}dim & class{\_}num & custom for number of classes in each dataset \\
    & n{\_}layers & 2 & number of dense layers \\
    & activation & relu & activation function \\
    \hline
\end{tabular}
\label{tab:hyperparameters}
\end{table}
\section{Expanded results}
\label{sec:expanded_results}
Modeling results for top-3 rankings are included below in Table \ref{tab:results_top-3}. It should be noted that since the ``CO (g)" category in the PKR dataset is a binary class (either yes or no), the top-3 accuracy will always be 1. This category is therefore excluded from AER calculations for this section.  
{\renewcommand{\arraystretch}{1.1}
\begin{table}[H]
\caption{Summary of top-3 ranking accuracies for all architectures across the four datasets.}
\vspace{0.1in}
\centering
\begin{tabular}{ c | c c c c c c c c c }
    \hline
    reaction & category & dummy & NFP & GGNN & MPNN & Weave & R-GAT & R-GCN & RS-GCN \\
    \hline
    \multirow{6}{3em}{\centering Suzuki} & \textbf{AER} & - & 0.3615 & 0.3491 & 0.0451 & 0.0847 & 0.2641 & \textbf{0.4936} & 0.2732 \\[1mm]
    & metal & 0.6744 & 0.8198 & 0.7935 & 0.7298 & 0.7388 & 0.7792 & \textbf{0.8482} & 0.7701 \\
    & ligand & 0.9269 & 0.9542 & 0.9555 & 0.9292 & 0.9351 & 0.9474 & \textbf{0.9644} & 0.9524 \\
    & base & 0.7344 & 0.7795 & 0.7693 & 0.7337 & 0.7366 & 0.7603 & \textbf{0.8123} & 0.7564 \\
    & solvent & 0.8013 & 0.8484 & 0.8430 & 0.7948 & 0.8055 & 0.8265 & \textbf{0.8836} & 0.8169 \\
    & additive & 0.9771 & 0.9904 & 0.9919 & 0.9784 & 0.9790 & 0.9884 & \textbf{0.9934} & 0.9899 \\[1mm]
    \hline
    \multirow{6}{3em}{\centering C–N} & \textbf{AER} & - & 0.4615 & 0.5240 & 0.0647 & 0.2077 & 0.3802 & \textbf{0.5391} & 0.3785 \\[1mm]
    & metal & 0.6526 & 0.8170 & 0.8392 & 0.6795 & 0.7393 & 0.7981 & \textbf{0.8479} & 0.7734 \\
    & ligand & 0.6647 & 0.8222 & 0.8532 & 0.6934 & 0.7203 & 0.7970 & \textbf{0.8605} & 0.7992 \\
    & base & 0.6400 & 0.8142 & 0.8326 & 0.6827 & 0.7360 & 0.7858 & \textbf{0.8452} & 0.7964 \\
    & solvent & 0.5677 & 0.7532 & 0.7847 & 0.5885 & 0.6538 & 0.7211 & \textbf{0.7973} & 0.7129 \\
    & additive & 0.9156 & 0.9537 & \textbf{0.9564} & 0.9151 & 0.9288 & 0.9433 & 0.9534 & 0.9471 \\[1mm]
    \hline
    \multirow{6}{3em}{\centering Negishi} & \textbf{AER} & - & 0.6503 & \textbf{0.6722} & 0.0896 & 0.2590 & 0.3598 & 0.6590 & 0.5148 \\[1mm]
    & metal & 0.5008 & 0.8054 & \textbf{0.8485} & 0.5072 & 0.6045 & 0.6715 & 0.8086 & 0.7512 \\
    & ligand & 0.8549 & \textbf{0.9601} & 0.9506 & 0.8724 & 0.8947 & 0.9187 & 0.9522 & 0.9474 \\
    & temperature & 0.5885 & 0.8262 & \textbf{0.8740} & 0.6619 & 0.7624 & 0.7608 & 0.8517 & 0.8086 \\
    & solvent & 0.8788 & 0.9522 & \textbf{0.9569} & 0.8852 & 0.9059 & 0.9171 & 0.9537 & 0.9394 \\
    & additive & 0.9043 & 0.9745 & 0.9681 & 0.9123 & 0.9203 & 0.9314 & \textbf{0.9761} & 0.9426 \\[1mm]
    \hline
    \multirow{9}{3em}{\centering PKR} & \textbf{AER} & - & 0.5957 & \textbf{0.6861} & 0.2695 & 0.3336 & 0.2947 & 0.6844 & 0.5063 \\[1mm]
    & metal & 0.7132 & 0.8604 & 0.8717 & 0.7849 & 0.8302 & 0.8189 & \textbf{0.9057} & 0.8604 \\
    & ligand & 0.9019 & \textbf{0.9887} & 0.9849 & 0.9811 & 0.9736 & 0.9736 & 0.9849 & \textbf{0.9887} \\
    & temperature & 0.5962 & 0.8038 & \textbf{0.8792} & 0.6415 & 0.6981 & 0.6604 & 0.8528 & 0.7509 \\
    & solvent & 0.5925 & 0.8340 & \textbf{0.8981} & 0.6981 & 0.7472 & 0.6981 & 0.8679 & 0.8226 \\
    & activator & 0.8830 & 0.9660 & 0.9698 & 0.8755 & 0.8906 & 0.8792 & \textbf{0.9774} & 0.9283 \\
    & CO (g) & 1.0000 & 1.0000 & 1.0000 & 1.0000 & 1.0000 & 1.0000 & 1.0000 & 1.0000 \\
    & additive & 0.9321 & 0.9698 & \textbf{0.9736} & 0.9472 & 0.9660 & 0.9509 & 0.9698 & \textbf{0.9736} \\
    & pressure & 0.9623 & 0.9774 & \textbf{0.9849} & 0.9736 & 0.9623 & 0.9736 & \textbf{0.9849} & 0.9698 \\[1mm]
    \hline
\end{tabular}
\label{tab:results_top-3}
\end{table}
}
\section{Exploratory data analysis (EDA)}
\label{sec:EDA}
\par A statistical analysis of the chemical space and reaction dictionary was conducted for each dataset. The analysis included distributions of reaction yields, categorical label frequencies, and the following 16 molecular descriptors of reaction products calculated with RDKit \citep{open-source_rdkit_2006}:
\begin{enumerate}
    \item MolWt = molecular weight (g/mol)
    \item MolLogP = molecular logP (lipophilicity measure)
    \item TPSA = topological polar surface area ($\si{\angstrom}^2$)
    \item HeavyAtomCount = number of non-H atoms
    \item NumHeteroatoms = number of heteroatoms (non-H or C)
    \item NumValenceElectrons = number of valence electrons
    \item NumHAcceptors = number of hydrogen-bond acceptors
    \item NumHDonors = number of hydrogen-bond donors
    \item NumRotatableBonds = number of rotatable bonds
    \item RingCount = number of rings
    \item NumAromHeterocycles = number of aromatic heterocycles (aromatic rings with at least one non-C atom)
    \item NumAromCarbocycles = number of aromatic carbocycles (aromatic rings made entirely of C atoms)
    \item NumSatHeterocycles = number of saturated heterocycles (saturated rings with at least one non-C atom)
    \item NumSatCarbocycles = number of saturated carbocycles (saturated rings made entirely of C atoms)
    \item FractionCSP3 = fraction of atoms $\textrm{sp}^3$-hybridized
    \item QED = quantitative estimation of drug-likeness
\end{enumerate}
\par Note: The descriptor names above are not all exactly as written in their respective RDKit functions. For the figures, the property distributions were truncated at the 1st and 99th percentile in each dataset analysis to avoid sparsity. The full span of the distributions is reflected in the summary tables. For full code, see the EDA jupyter notebooks in the associated GitHub repository.
\par \textbf{Suzuki dataset}.
\begin{table}[H]
\caption{Summary of product molecular properties in Suzuki dataset.}
\vspace{0.1in}
\respectpercent
\begin{tabular}{c c c c c c c c}
\hline
\begin{filecontents*}{suzuki_stats.csv}
,Yield,MolWt,MolLogP,TPSA,HeavyAtomCount,NumHeteroatoms,NumValenceElectrons,NumHAcceptors,NumHDonors,NumRotatableBonds,RingCount,NumAromHeterocycles,NumAromCarbocycles,NumSatHeterocycles,NumSatCarbocycles,FractionCSP3,QED
count,145413,144972,144972,144972,144972,144972,144972,144972,144972,144972,144972,144972,144972,144972,144972,144972,144972
mean,0.682,352.072,4.704,49.805,25.323,4.938,129.042,3.622,0.525,3.968,3.472,1.028,1.976,0.171,0.097,0.188,0.544
std,0.237,146.421,2.258,36.775,10.435,3.421,54.925,2.589,0.832,3.483,1.639,1.05,1.248,0.45,0.394,0.166,0.201
min,0,82.106,-5.969,0,6,0,32,0,0,0,0,0,0,0,0,0,0.008
25%,0.53,233.227,3.354,20.23,17,2,84,2,0,2,2,0,1,0,0,0.067,0.396
50%,0.73,331.757,4.142,43.6,24,4,120,3,0,3,3,1,2,0,0,0.143,0.591
75%,0.88,438.444,5.519,72.45,31,7,160,5,1,5,4,2,2,0,0,0.3,0.695
max,1,2339.16,46.845,472.09,164,86,952,39,16,115,11,8,9,7,5,1,0.948
\end{filecontents*}
\csvreader[mystyle]{suzuki_stats.csv}{}{\csvcoli & \csvcolii & \csvcoliii & \csvcoliv & \csvcolv & \csvcolvi & \csvcolvii & \csvcolviii}
\end{tabular}
\begin{tabular}{c c c c c c}
\\
\hline
\csvreader[mystyle]{suzuki_stats.csv}{}{\csvcoli & \csvcolix & \csvcolx & \csvcolxi & \csvcolxii & \csvcolxiii}
\end{tabular}
\begin{tabular}{c c c c c c}
\\
\hline
\csvreader[mystyle]{suzuki_stats.csv}{}{\csvcoli & \csvcolxiv & \csvcolxv & \csvcolxvi & \csvcolxvii & \csvcolxviii}
\end{tabular}
\label{tab:suzuki_props}
\end{table}
\begin{figure}[H]
    \centering
    \caption{Distribution of reaction yields in Suzuki dataset.}
    \vspace{0.1in}
    \includegraphics[width = 0.45\textwidth]{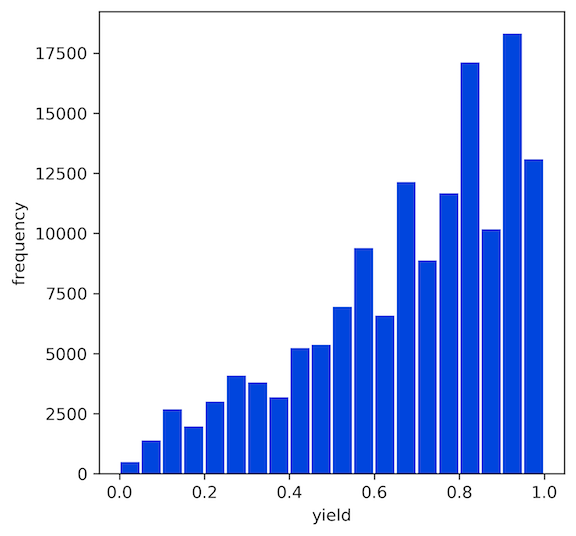}
    \label{fig:suzuki_yield}
\end{figure}
\begin{figure}[H]
    \centering
    \caption{Distribution of each molecular descriptor in Suzuki dataset products.}
    \vspace{0.1in}
    \includegraphics[width = 0.95\textwidth]{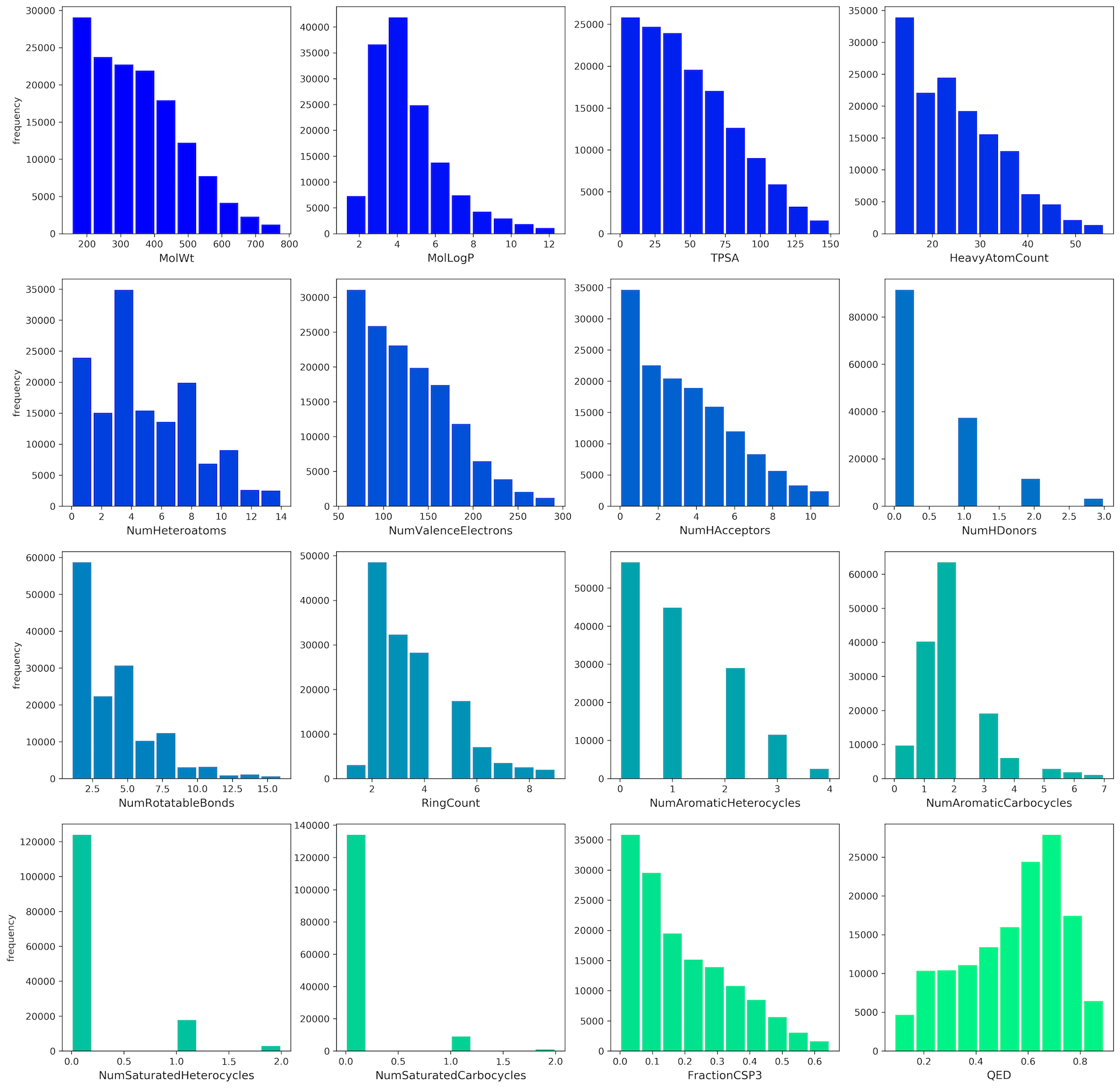}
    \label{fig:suzuki_props}
\end{figure}
\begin{figure}[H]
    \centering
    \caption{Distribution of dictionary bin frequencies in Suzuki dataset.}
    \vspace{0.1in}
    \includegraphics[width = 0.95\textwidth]{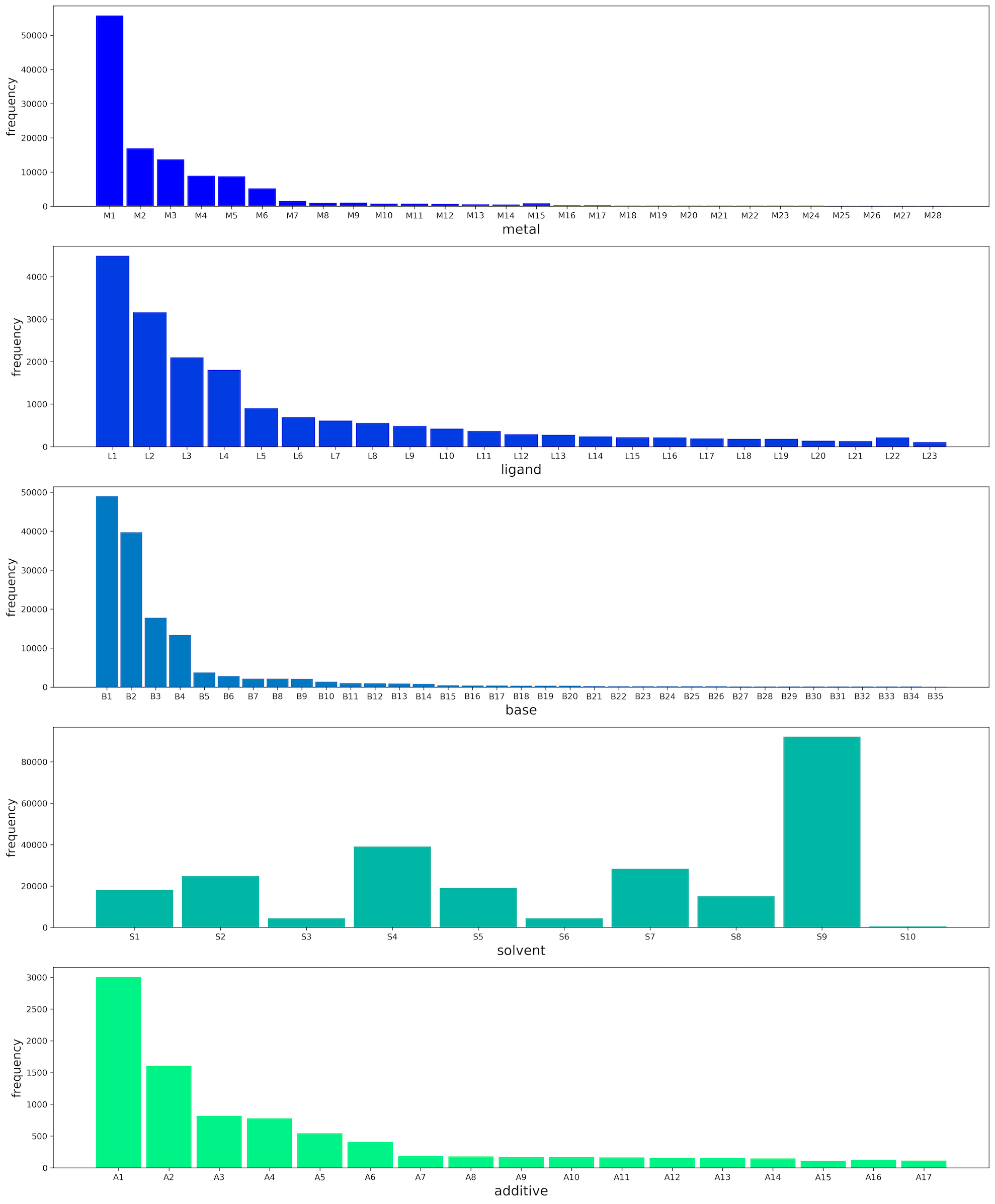}
    \label{fig:suzuki_dict}
\end{figure}
\par \textbf{C–N dataset}.
\begin{table}[H]
\caption{Summary of product molecular properties in C–N dataset.}
\vspace{0.1in}
\respectpercent
\begin{tabular}{c c c c c c c c}
\hline
\begin{filecontents*}{cn_stats.csv}
,Yield,MolWt,MolLogP,TPSA,HeavyAtomCount,NumHeteroatoms,NumValenceElectrons,NumHAcceptors,NumHDonors,NumRotatableBonds,RingCount,NumAromHeterocycles,NumAromCarbocycles,NumSatHeterocycles,NumSatCarbocycles,FractionCSP3,QED
count,36519,36504,36504,36504,36504,36504,36504,36504,36504,36504,36504,36504,36504,36504,36504,36504,36504
mean,0.699,310.3,4.281,41.023,22.272,4.237,114.149,3.128,0.673,3.598,3.022,0.452,2.093,0.292,0.067,0.225,0.603
std,0.229,123.812,2.279,32.613,8.887,2.82,44.884,2.018,0.771,2.804,1.643,0.691,1.292,0.53,0.361,0.191,0.197
min,0,107.156,-1.441,3.01,8,1,42,1,0,1,1,0,1,0,0,0,0.015
25%,0.57,216.24,2.87,12.47,16,2,82,2,0,2,2,0,1,0,0,0.067,0.481
50%,0.75,279.325,3.763,33.2,20,4,102,3,1,3,3,0,2,0,0,0.182,0.647
75%,0.88,376.4,5.166,58.44,27,6,136,4,1,5,4,1,2,1,0,0.4,0.757
max,1,2527.86,30.826,355.67,172,40,960,39,8,114,12,7,9,4,5,0.917,0.948
\end{filecontents*}
\csvreader[mystyle]{cn_stats.csv}{}{\csvcoli & \csvcolii & \csvcoliii & \csvcoliv & \csvcolv & \csvcolvi & \csvcolvii & \csvcolviii}
\end{tabular}
\begin{tabular}{c c c c c c}
\\
\hline
\csvreader[mystyle]{cn_stats.csv}{}{\csvcoli & \csvcolix & \csvcolx & \csvcolxi & \csvcolxii & \csvcolxiii}
\end{tabular}
\begin{tabular}{c c c c c c}
\\
\hline
\csvreader[mystyle]{cn_stats.csv}{}{\csvcoli & \csvcolxiv & \csvcolxv & \csvcolxvi & \csvcolxvii & \csvcolxviii}
\end{tabular}
\label{tab:cn_props}
\end{table}
\begin{figure}[H]
    \centering
    \caption{Distribution of reaction yields in C–N dataset.}
    \vspace{0.1in}
    \includegraphics[width = 0.45\textwidth]{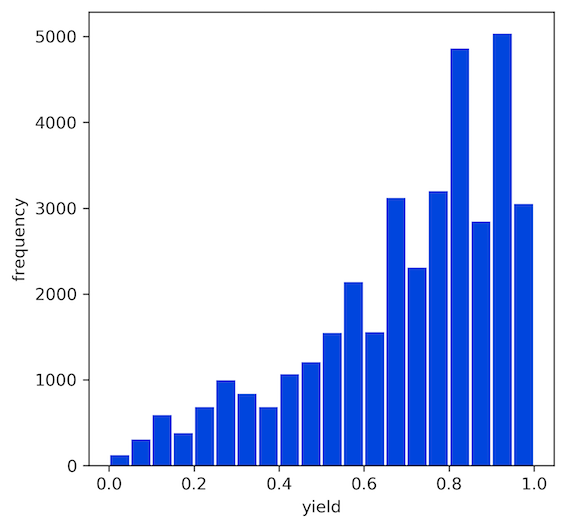}
    \label{fig:cn_yield}
\end{figure}
\begin{figure}[H]
    \centering
    \caption{Distribution of each molecular descriptor in C–N dataset products.}
    \vspace{0.1in}
    \includegraphics[width = 0.95\textwidth]{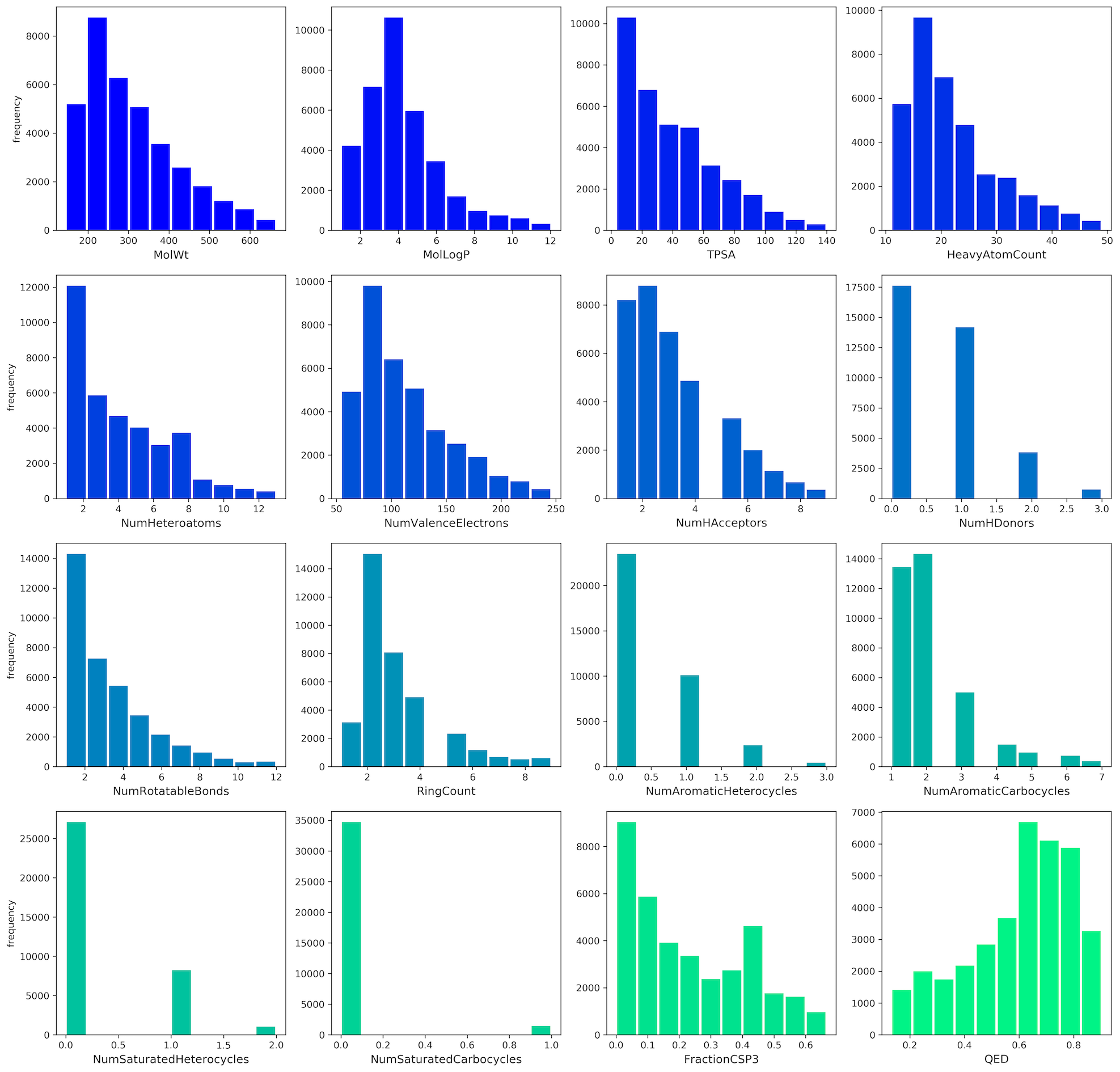}
    \label{fig:cn_props}
\end{figure}
\begin{figure}[H]
    \centering
    \caption{Distribution of dictionary bin frequencies in C–N dataset.}
    \vspace{0.1in}
    \includegraphics[width = 0.95\textwidth]{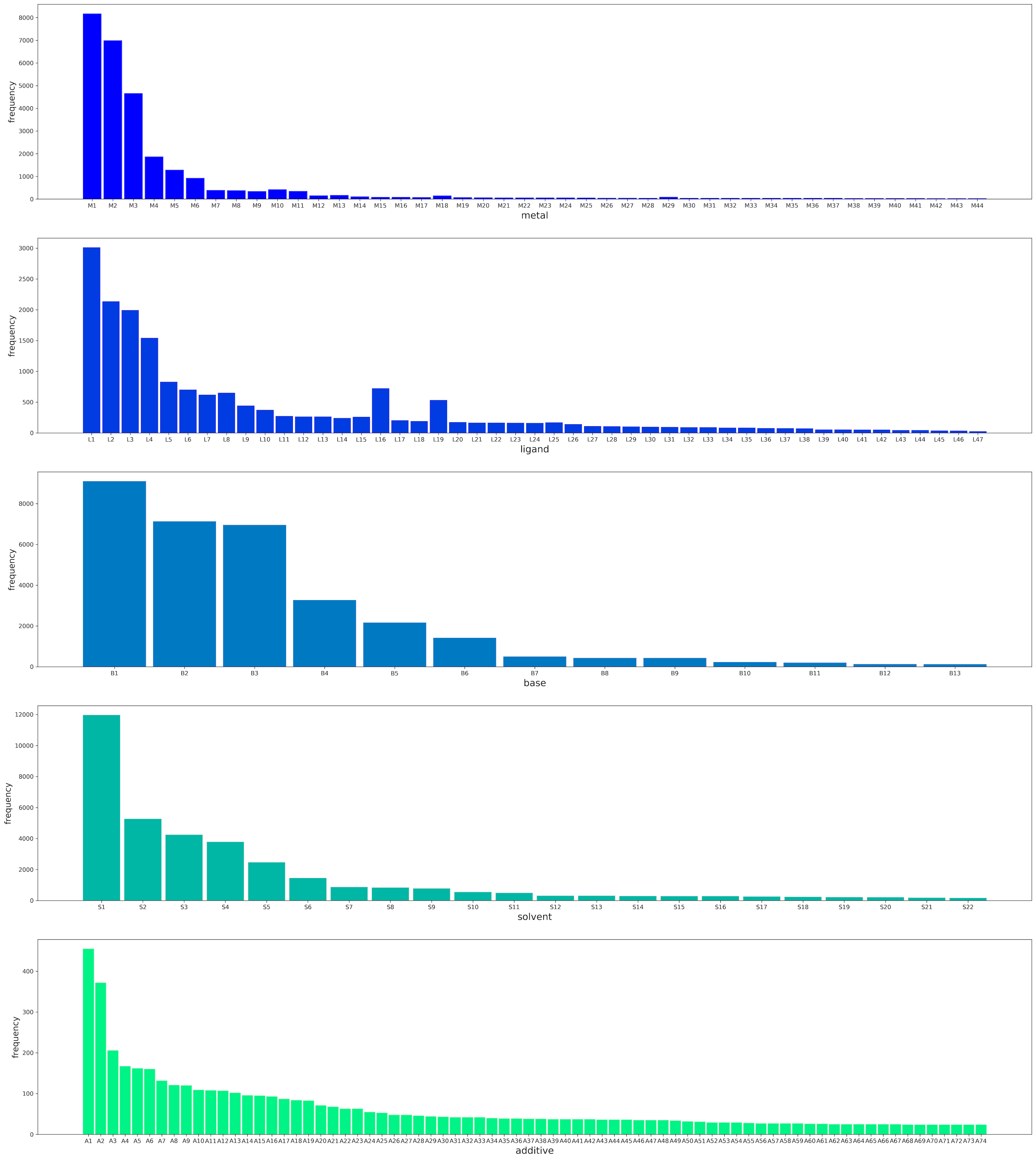}
    \label{fig:cn_dict}
\end{figure}
\par \textbf{Negishi dataset}.
\begin{table}[H]
\caption{Summary of product molecular properties in Negishi dataset.}
\vspace{0.1in}
\respectpercent
\begin{tabular}{c c c c c c c c}
\hline
\begin{filecontents*}{negishi_stats.csv}
,Yield,MolWt,MolLogP,TPSA,HeavyAtomCount,NumHeteroatoms,NumValenceElectrons,NumHAcceptors,NumHDonors,NumRotatableBonds,RingCount,NumAromHeterocycles,NumAromCarbocycles,NumSatHeterocycles,NumSatCarbocycles,FractionCSP3,QED
count,6391,6383,6383,6383,6383,6383,6383,6383,6383,6383,6383,6383,6383,6383,6383,6383,6383
mean,0.711,288.295,3.888,39.464,20.4,3.92,106.51,2.867,0.246,3.599,2.344,0.68,1.224,0.134,0.149,0.297,0.615
std,0.219,111.961,1.677,28.719,7.858,2.679,42.459,1.992,0.53,2.823,1.238,0.824,0.915,0.392,0.494,0.213,0.17
min,0.005,82.102,-1.691,0,6,0,32,0,0,0,0,0,0,0,0,0,0.036
25%,0.59,209.248,2.815,18.46,15,2,76,1,0,2,2,0,1,0,0,0.118,0.529
50%,0.76,263.243,3.639,35.53,19,3,98,3,0,3,2,0,1,0,0,0.273,0.649
75%,0.889,339.369,4.651,55.14,24,5,126,4,0,5,3,1,2,0,0,0.45,0.741
max,1,1209.1,17.401,213.58,80,28,456,17,6,31,9,5,7,4,4,0.943,0.946
\end{filecontents*}
\csvreader[mystyle]{negishi_stats.csv}{}{\csvcoli & \csvcolii & \csvcoliii & \csvcoliv & \csvcolv & \csvcolvi & \csvcolvii & \csvcolviii}
\end{tabular}
\begin{tabular}{c c c c c c}
\\
\hline
\csvreader[mystyle]{negishi_stats.csv}{}{\csvcoli & \csvcolix & \csvcolx & \csvcolxi & \csvcolxii & \csvcolxiii}
\end{tabular}
\begin{tabular}{c c c c c c}
\\
\hline
\csvreader[mystyle]{negishi_stats.csv}{}{\csvcoli & \csvcolxiv & \csvcolxv & \csvcolxvi & \csvcolxvii & \csvcolxviii}
\end{tabular}
\label{tab:negishi_props}
\end{table}
\begin{figure}[H]
    \centering
    \caption{Distribution of reaction yields in Negishi dataset.}
    \vspace{0.1in}
    \includegraphics[width = 0.45\textwidth]{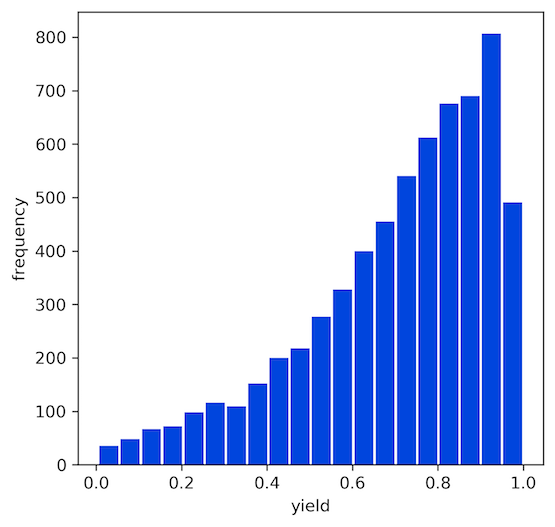}
    \label{fig:negishi_yield}
\end{figure}
\begin{figure}[H]
    \centering
    \caption{Distribution of each molecular descriptor in Negishi dataset products.}
    \vspace{0.1in}
    \includegraphics[width = 0.95\textwidth]{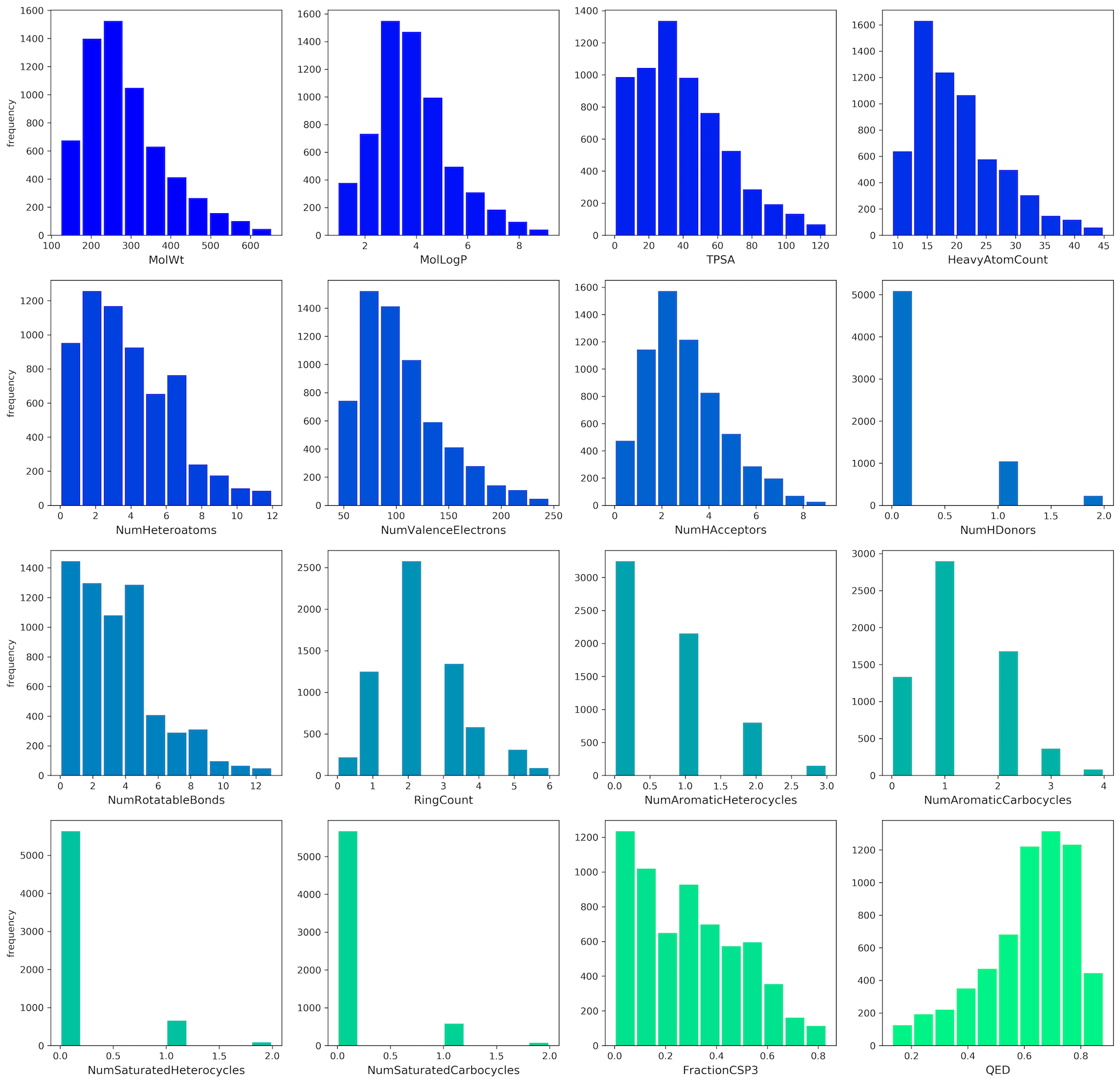}
    \label{fig:negishi_props}
\end{figure}
\begin{figure}[H]
    \centering
    \caption{Distribution of dictionary bin frequencies in Negishi dataset.}
    \vspace{0.1in}
    \includegraphics[width = 0.95\textwidth]{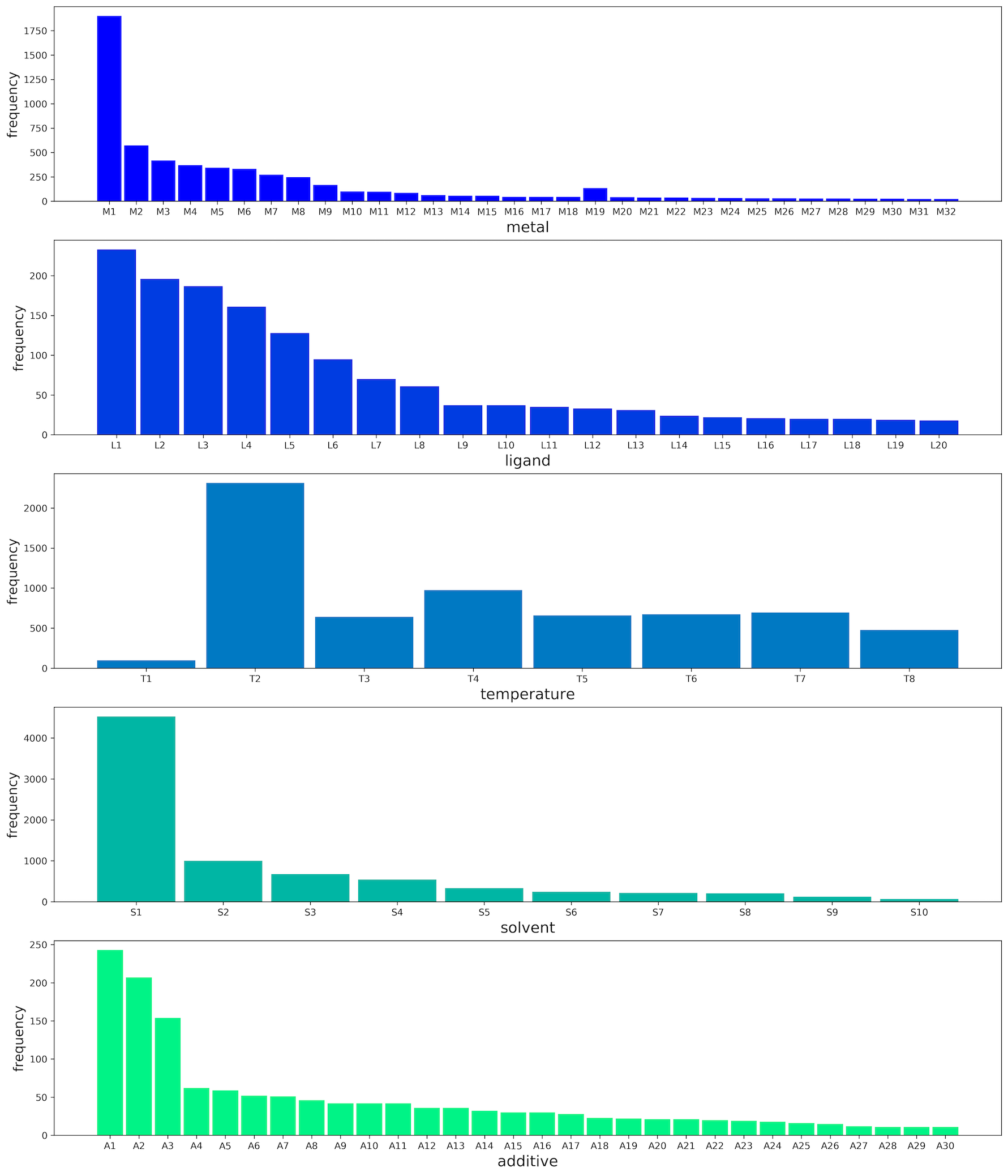}
    \label{fig:negishi_dict}
\end{figure}
\par \textbf{PKR dataset}.
\begin{table}[H]
\caption{Summary of product molecular properties in PKR dataset.}
\vspace{0.1in}
\respectpercent
\begin{tabular}{c c c c c c c c}
\hline
\begin{filecontents*}{pkr_stats.csv}
,Yield,MolWt,MolLogP,TPSA,HeavyAtomCount,NumHeteroatoms,NumValenceElectrons,NumHAcceptors,NumHDonors,NumRotatableBonds,RingCount,NumAromHeterocycles,NumAromCarbocycles,NumSatHeterocycles,NumSatCarbocycles,FractionCSP3,QED
count,2749,2746,2746,2746,2746,2746,2746,2746,2746,2746,2746,2746,2746,2746,2746,2746,2746
mean,0.662,312.949,3.399,45.976,22.334,3.997,117.857,3.146,0.111,3.038,3.26,0.047,0.959,0.425,0.538,0.472,0.636
std,0.218,94.145,1.705,23.243,6.508,1.992,34.337,1.643,0.347,2.357,1.085,0.216,0.983,0.64,0.713,0.203,0.153
min,0.03,122.167,-1.182,0,9,0,48,0,0,0,1,0,0,0,0,0,0.051
25%,0.51,240.302,2.096,26.3,17,2,92,2,0,1,3,0,0,0,0,0.308,0.555
50%,0.69,303.393,3.125,44.76,22,4,114,3,0,3,3,0,1,0,0,0.45,0.663
75%,0.84,364.4,4.374,63.68,26,5,136,4,0,4,4,0,1,1,1,0.643,0.747
max,1,905.243,11.217,151.35,64,13,340,13,3,24,8,2,5,4,4,0.929,0.924
\end{filecontents*}
\csvreader[mystyle]{pkr_stats.csv}{}{\csvcoli & \csvcolii & \csvcoliii & \csvcoliv & \csvcolv & \csvcolvi & \csvcolvii & \csvcolviii}
\end{tabular}
\begin{tabular}{c c c c c c}
\\
\hline
\csvreader[mystyle]{pkr_stats.csv}{}{\csvcoli & \csvcolix & \csvcolx & \csvcolxi & \csvcolxii & \csvcolxiii}
\end{tabular}
\begin{tabular}{c c c c c c}
\\
\hline
\csvreader[mystyle]{pkr_stats.csv}{}{\csvcoli & \csvcolxiv & \csvcolxv & \csvcolxvi & \csvcolxvii & \csvcolxviii}
\end{tabular}
\label{tab:pkr_props}
\end{table}
\begin{figure}[H]
    \centering
    \caption{Distribution of reaction yields in PKR dataset.}
    \vspace{0.1in}
    \includegraphics[width = 0.45\textwidth]{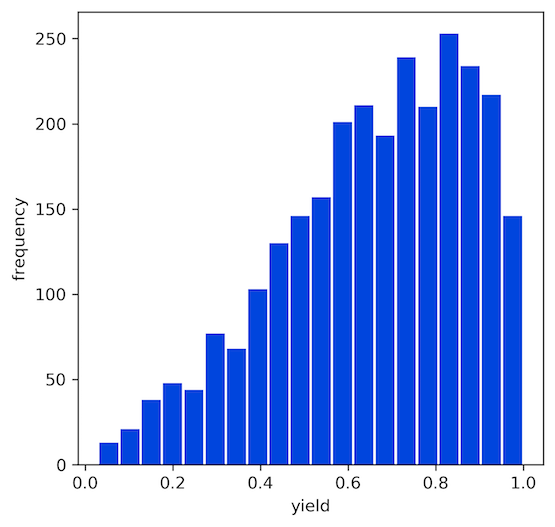}
    \label{fig:pkr_yield}
\end{figure}
\begin{figure}[H]
    \centering
    \caption{Distribution of each molecular descriptor in PKR dataset products.}
    \vspace{0.1in}
    \includegraphics[width = 0.95\textwidth]{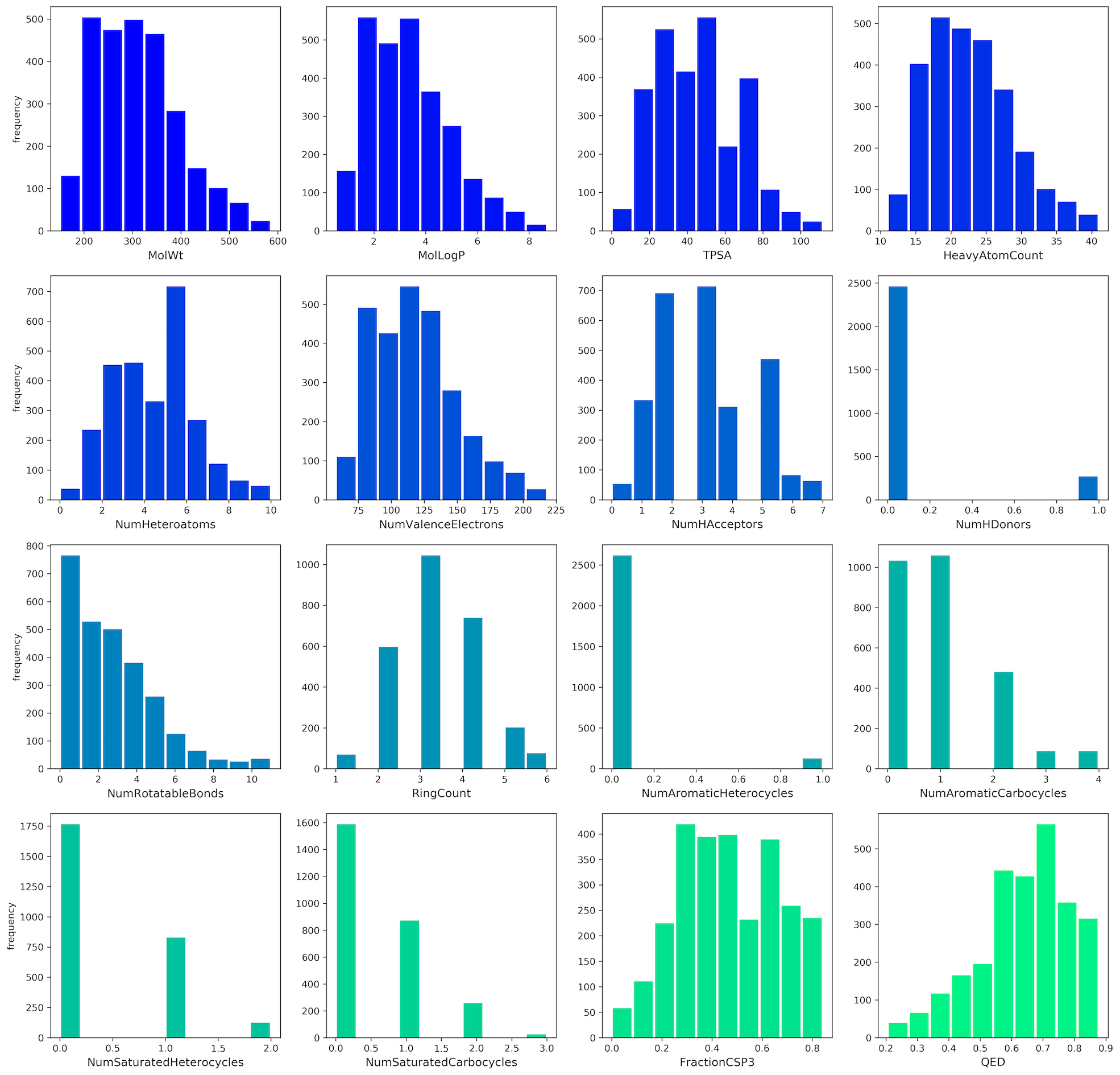}
    \label{fig:pkr_props}
\end{figure}
\begin{figure}[H]
    \centering
    \caption{Distribution of dictionary bin frequencies in PKR dataset.}
    \vspace{0.1in}
    \includegraphics[width = 0.95\textwidth]{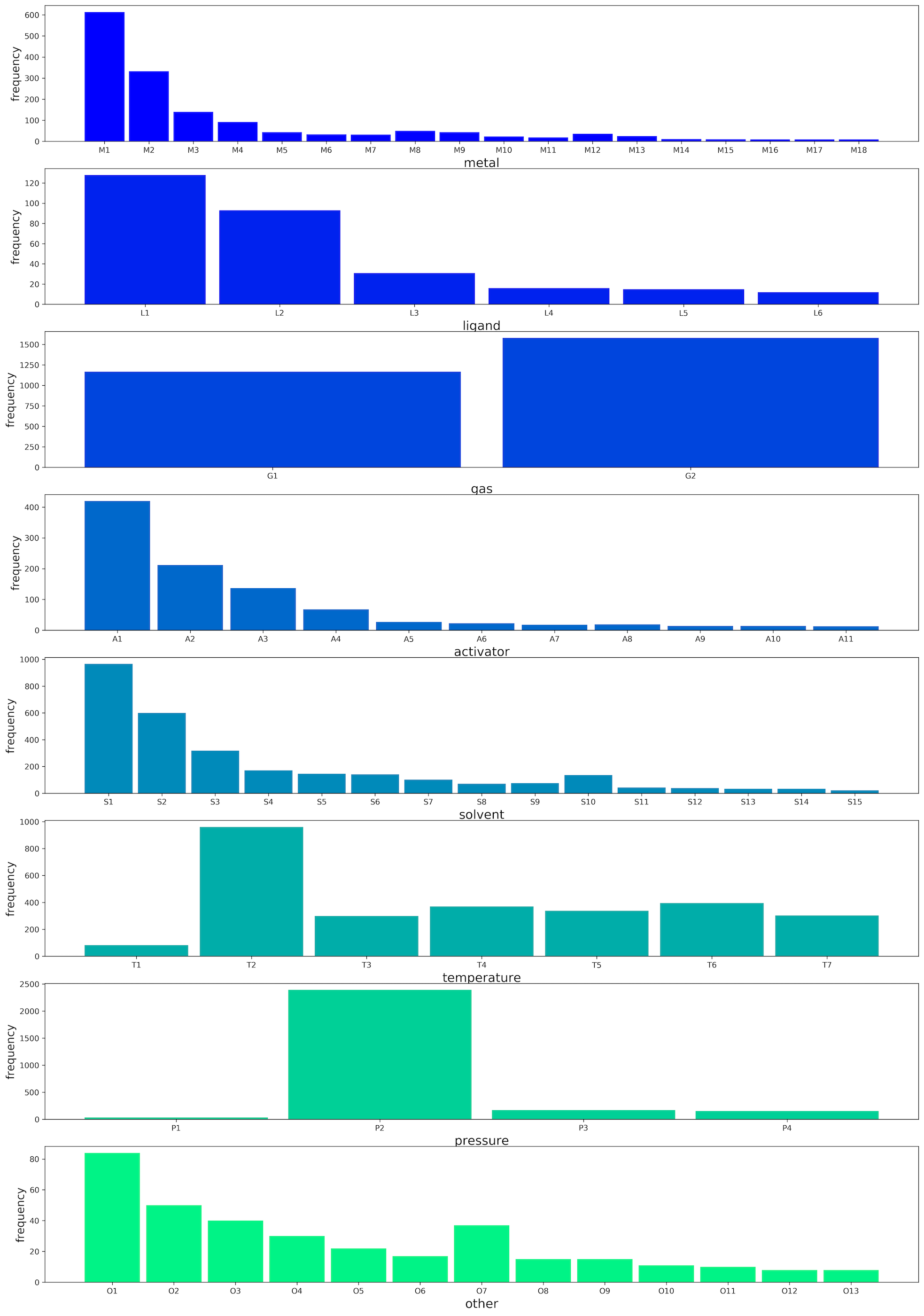}
    \label{fig:pkr_dict}
\end{figure}

\end{document}